\documentclass[11pt]{scaleai-paper}

\usepackage{amsmath}
\usepackage{amsfonts}
\usepackage{amssymb}
\usepackage{amsthm}
\usepackage{booktabs}
\usepackage{tabularx}
\usepackage{tabulary}
\usepackage{multirow}
\usepackage{subcaption}
\usepackage{float}
\usepackage[square,numbers,sort&compress]{natbib}
\usepackage{xspace}
\usepackage{url}
\usepackage[colorlinks=true,linkcolor=scaleLink,citecolor=scaleLink,urlcolor=scaleLink]{hyperref}
\usepackage[capitalise,nameinlink]{cleveref}

\usepackage{graphicx}
\usepackage{pgfplots}
\pgfplotsset{compat=1.18}
\newcolumntype{Y}{>{\RaggedRight\arraybackslash}X}
\let\svthefootnote\thefootnote
\newcommand\freefootnote[1]{%
  \let\thefootnote\relax%
  \footnotetext{#1}%
  \let\thefootnote\svthefootnote%
}

\usepackage{xcolor}

\newcommand{\benchmarkname}{\textsc{CliniCARE-Bench}}

\contact{\texttt{veronica.chatrath@scale.com, yuan.xue@scale.com} \quad | \quad \url{https://scale.com/research}}

\title{CliniCARE-Bench: Clinical Calibrated Audit of Medical Reasoning in EHR}

\author[1]{Veronica Chatrath\textsuperscript{*}}
\author[1]{Bryan Zhu\textsuperscript{*}}
\author[1]{George Pu}
\author[1]{Jingxuan Fan}
\author[1]{Apaar Shanker}
\author[1]{Varun Ursekar}
\author[1]{Anahita Sharma}
\author[1]{Jason Qin}
\author[1,2]{Keqi Han}
\author[1]{Soham Dinesh Tiwari}
\author[1]{Soham Dan}
\author[1]{Vijay Kalmath}
\author[1]{Yuan (Christy) Li}
\author[1]{Daniel Yue Zhang}
\author[1,4]{Chenguang Wang}
\author[1]{Zainab Doctor}
\author[1,3]{Zhijun Yin}
\author[5, 6, 7]{Nigam H. Shah}
\author[1]{Yuan Xue}

\affil[1]{Scale AI}
\affil[2]{Emory University}
\affil[3]{Vanderbilt University Medical Center}
\affil[4]{University of California, Santa Cruz}
\affil[5]{Department of Medicine, Stanford School of Medicine}
\affil[6]{Technology and Digital Solutions, Stanford Health Care}
\affil[7]{Clinical Excellence Research Center, Stanford School of Medicine}
\affil[*]{Co-first authors.}

\begin{document}


\maketitle

\begin{abstract}
Large language models perform strongly on medical knowledge benchmarks, but reliable clinical deployment requires agents to conduct defensible investigations over heterogeneous, longitudinal records. Agents must determine what evidence is needed, retrieve and reconcile structured and free-text data, ground conclusions in verifiable evidence, and defer cases that cannot be resolved reliably. We introduce \benchmarkname{} (\textbf{Clini}cal \textbf{C}alibrated \textbf{A}udit of Medical \textbf{R}easoning in \textbf{E}HR), a deployment-oriented benchmark for retrospective clinical audit: 25 clinician-validated scenarios instantiated as 750 patient-specific cases over real-patient-derived MIMIC-IV data. Systems investigate each case through a governed, logged tool environment for record retrieval, computation, and policy access, and return one of four verdicts---\textit{Yes}, \textit{No}, \textit{Indeterminate: Lack of Data}, or \textit{Indeterminate: Medically Ambiguous}---the last two separating missing evidence from residual medical ambiguity. Beyond verdict accuracy, we score patient-evidence and policy grounding, process adherence, calibrated abstention, reliability, and efficiency against case-level reference verdicts produced by independent multi-model adjudication and calibrated against Clinical Board review. As every retrieval, computation, and report is replayable, the investigation and adjudication trace is itself inspectable and scorable. To our knowledge, \benchmarkname{} is the first deployment-oriented clinical-agent benchmark to jointly evaluate real longitudinal EHR investigation, claim-level evidence grounding, governing-policy use, process adherence, and calibrated abstention within a common patient-level adjudication framework. In the evaluation of 16 agentic systems, four-way accuracy spans 65.3--76.1\%. However, raw accuracy overstates the quality of the underlying investigation. Defect-free accuracy, which credits a verdict only when it is correct and free of prohibited shortcuts, is 4.8--14.8 percentage points lower and reorders the leaderboard. We will release the scenarios and evaluation code soon.
\end{abstract}

\section{Introduction}

Many high-value clinical AI tasks are not self-contained questions or prospective treatment decisions but retrospective investigations of the patient record. Was a protocol followed? Did a clinical event occur within a given window? Does the documentation support a quality or operational conclusion? Answering any of these means working through a heterogeneous, longitudinal electronic health record (EHR), deciding what evidence is needed, retrieving it across structured data and free-text notes, reconstructing the timeline, reconciling conflicting sources, applying the governing clinical standard, and recognizing when the record simply cannot settle the question. This is evidence-grounded clinical audit and adjudication, not medical question answering.

Foundation models now match or exceed expert scores on medical knowledge exams \cite{singhal2023llm,nori2023gpt4medical}, but exam performance is a poor proxy for this investigation and adjudication process. Existing clinical benchmarks each capture part of that process. Knowledge suites probe internalized concepts with no environment to act in \cite{soskinhicks2026healthbenchpro}. Interactive simulators drive sequential diagnosis on synthetic or partly simulated substrates \cite{schmidgall2024agentclinic,luo2026clinicalenvsim}. EHR-grounded frameworks span heterogeneous cases \cite{bedi2025medhelm}, expert instructions \cite{fleming2024medalign}, and multi-year timelines \cite{wornow2023ehrshot}. Factuality benchmarks check whether generated statements are supported by the record \cite{chung2026verifact,munnangi2024factehr}. Three safety-critical capabilities, however, remain poorly tested: \emph{negative reasoning} (confirming a condition was considered and ruled out), \emph{deep longitudinal synthesis} across multi-year records \cite{cui2025timer}, and \emph{calibrated abstention} that separates missing evidence from genuine medical ambiguity and defers rather than guessing when the record is incomplete or contradictory \cite{watanabe2026clindet, machcha2026medabstain}. More fundamentally, few benchmarks treat the \emph{defensibility} of an adjudication as the unit of evaluation: whether the agent gathered the necessary evidence, grounded its claims, applied the governing standard, and escalated when the record could not support a verdict. Deployment demands more than a correct final answer. It demands an evidence acquisition and adjudication process that is complete, faithful, reproducible, and calibrated.

\begin{figure}[ht!]
\centering
\includegraphics[width=\linewidth]{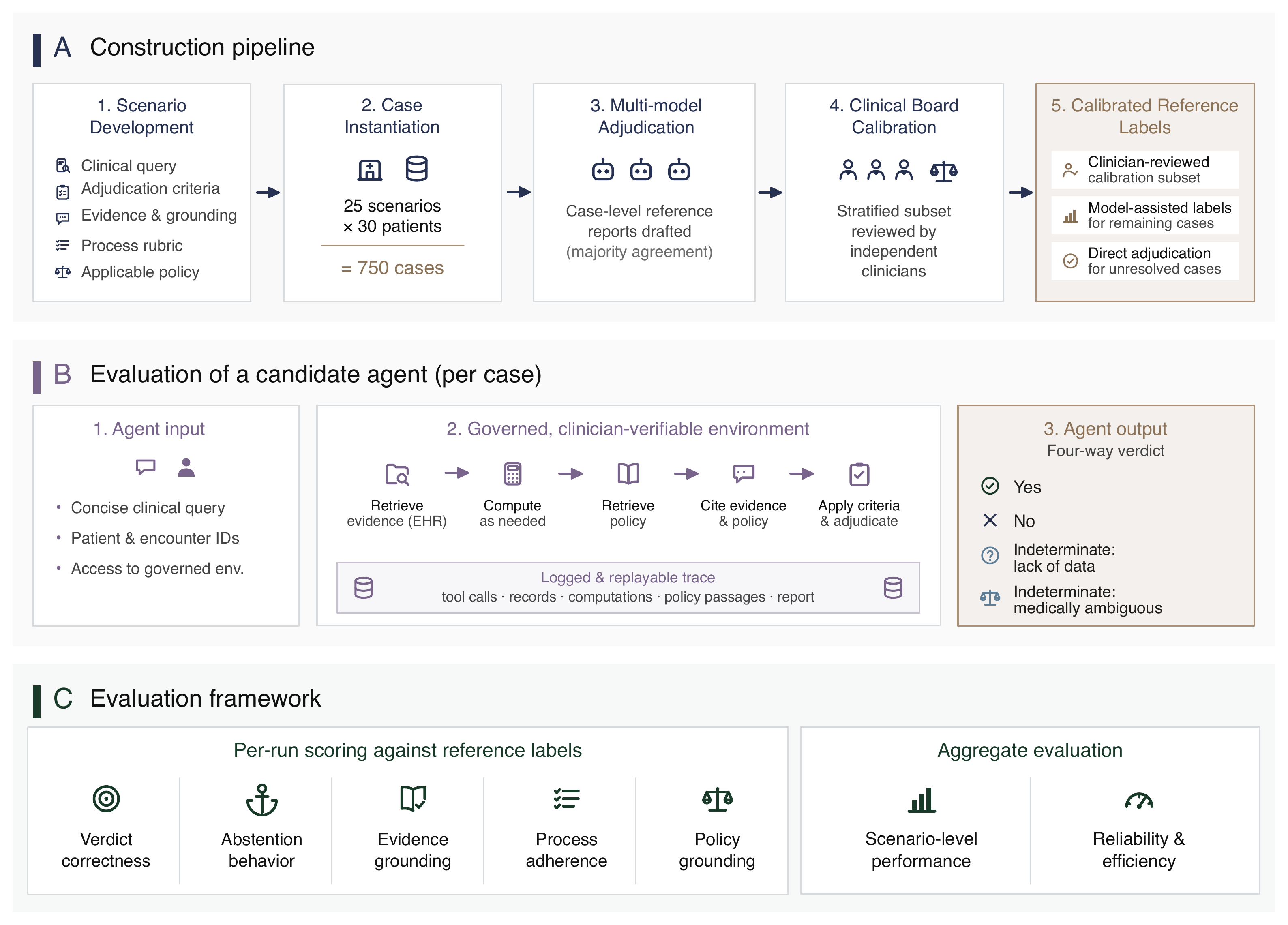}
\caption{\textbf{(A)} Clinician-authored scenarios are instantiated as patient-specific MIMIC-IV cases, independently adjudicated by a multi-model ensemble, and calibrated against Clinical Board review into case-level reference verdicts. \textbf{(B)} Per case, a system answers a clinical query within a governed environment for record retrieval, computation, and policy access, producing a cited four-way verdict and a replayable trace. \textbf{(C)} Runs are scored on verdict correctness, abstention, evidence grounding, process adherence, and policy grounding, then aggregated into scenario-level performance, reliability, efficiency, and cost.}
\label{fig:overview}
\end{figure}

We introduce \textbf{CliniCARE-Bench} (\textbf{Clini}cal \textbf{C}alibrated \textbf{A}udit of Medical \textbf{R}easoning in \textbf{E}HR), a suite of 25 clinician-authored, evidence-intensive scenarios instantiated as 750 patient cases over MIMIC-IV v3.1 \cite{physionet-mimic-iv} (Figure~\ref{fig:overview}). Rather than handing agents a preassembled context or raw database access, we place them in a reproducible runtime with governed, clinician-verifiable tools for retrieving structured records and notes, running explicit computations, and consulting a fixed corpus of governing policies, isolating query engineering artifacts while preserving the record's heterogeneity. Agents must combine medical knowledge, patient evidence, and authoritative standards such as KDIGO \cite{kdigo2012aki}, CMS SEP-1 \cite{SevereSepsisSeptic}, AABB \cite{carson2023rbc}, and AHA/ACC guidance \cite{heidenreich2022hf}.

Every case demands one of four verdicts: \textit{Yes}, \textit{No}, \textit{Indeterminate: Lack of Data}, or \textit{Indeterminate: Medically Ambiguous}. The two indeterminate classes separate cases that more evidence could resolve from those that need expert interpretation even after a full review. Reference verdicts come from a multi-model adjudication ensemble applying the clinician-validated specifications, calibrated against blinded Clinical Board review on a stratified subset. Abstention is thus built into the evaluation standard, not bolted on as a post hoc confidence threshold.

To our knowledge, \benchmarkname{} is the first \emph{deployment-oriented} clinical agent benchmark to score real longitudinal EHR investigation, claim-level evidence grounding, governing policy use, process adherence, and calibrated abstention \emph{jointly} within one patient-level adjudication framework. No benchmark in Table~\ref{tab:benchmark_positioning} combines all of these axes. It builds on the trace-level clinical auditing introduced by \citet{hager2024clinicaldecision} but broadens what is scored within the trace. Every retrieval and computation is logged, and every verdict must cite the patient evidence it rests on, together with any governing policy that applies. A defensible conclusion can then be told apart from a lucky guess or an invalid shortcut, without ever inspecting the model's private reasoning. Making this feasible at scale is a protocol for reference verdicts, calibrated against clinician review, that replaces exhaustive manual labeling.

Throughout the paper, we refer to the task-performing entity as the \emph{agent}, and to the complete evaluated configurations, including the model, agent harness, prompting or scaffolding, and available tools, as the \emph{system}. In an evaluation of 16 systems combining production agent harnesses with frontier models, as well as open-weight models evaluated on a common harness, even the best system resolves only 76.1\% of cases correctly. Every system also under-abstains: the over-commitment rate on cases whose reference verdict requires deferral exceeds the over-abstention rate on cases with definitive reference verdicts. These results suggest substantial remaining gaps toward reliable autonomous clinical audit.

We make four methodological contributions:

\begin{enumerate}
\item \textbf{A skill- and specialty-spanning scenario suite.} 25 clinician-developed, evidence-intensive scenarios instantiated as 750 patient cases over MIMIC-IV, spanning protocol and guideline audits, clinical event determination, and longitudinal status assessment across 14 medical specialties drawn from the American Board of Medical
Specialties (ABMS) taxonomy\footnote{\url{https://www.abms.org/member-boards/specialty-subspecialty-certificates/}} and ten reasoning capabilities, stress-testing safety-critical reasoning that existing benchmarks rarely evaluate.

\item \textbf{A governed, clinician-verifiable tool environment.} A reproducible execution surface over real longitudinal EHR data with patient-scoped retrieval, explicit computation, policy document access, and fully logged interactions, so performance reflects clinical investigation rather than query engineering.

\item \textbf{A grounded, process-aware evaluation framework.} A strict four-way verdict paired with claim-level evidence grounding, required evidence coverage, policy citation and adherence, and a weighted \textsc{must-do}/\textsc{must-not} process rubric rewarding faithful evidence use, appropriate abstention, and sound procedure, not correctness of the final answer alone.

\item \textbf{A scalable protocol for clinician-calibrated reference labeling.} Specifications authored and validated by clinicians, combined with independent multi-model case adjudication and blinded Clinical Board calibration, yielding reliable reference verdicts at a scale exhaustive manual review could not reach.

\end{enumerate}

\noindent\textbf{Key empirical finding: correct verdicts can mask defective investigations.}
Defect-free accuracy credits a verdict only when it is correct and the
investigation violates no prohibited \textsc{must-not} criterion. Across
the 16 systems, defect-free accuracy is 4.8--14.8 percentage points below
accuracy on the same report-present runs, and this correction changes the
ordering of systems. For the most affected system, roughly one in five otherwise-correct verdicts violates at least one prohibited shortcut criterion. This extends the outcome--process gap documented for medical LLMs \cite{bedi2025fidelity,natmed2026robustness} to agentic investigation over real longitudinal EHRs, and captures precisely the failure mode that CliniCARE-Bench is designed to expose.

Together these pieces form an open benchmark for whether clinical agents have the safety-critical capabilities that real-world audit and review work demands. Though instantiated in medicine, our design, including governed, auditable tool use, explicit adjudication criteria, traceable evidence, calibrated abstention, and scoring of the observable adjudication process, transfers to other high-stakes domains such as finance, cybersecurity, and law, where decisions must be transparent, reproducible, and verifiable. We will release the benchmark data, evaluation code, and reproducibility resources, for authorized MIMIC-IV users.
\section{Related Work} 

\subsection{Static Knowledge-Centric and Context-Provided Clinical Evaluation}

Early medical LLM benchmarks primarily assess whether models can apply medical knowledge and reasoning to predefined question-answering tasks. For example, MultiMedQA \cite{singhal2023llm} and related medical-examination benchmarks \cite{nori2023gpt4medical} combine professional examination questions, biomedical research questions, and consumer health queries. Although these benchmarks span diverse clinical domains, the information required to answer each question is generally contained in the prompt or encoded in the model’s parameters. More recent rubric-based benchmarks extend evaluation from closed-form questions to open-ended interactions. HealthBench \cite{arora2025healthbench} evaluates healthcare conversations using physician-authored criteria, while HealthBench Professional \cite{soskinhicks2026healthbenchpro} focuses on cases drawn from model–clinician interactions, including clinical consultation, documentation, and medical research. These benchmarks improve the realism of clinical application, but they primarily evaluate how models respond to information made available within a predefined interaction context rather than how they acquire evidence from a longitudinal patient record. 

Beyond conversational evaluation, another line of work grounds model assessment in patient-derived EHR data assembled into case-specific inputs or contexts. MedAlign \cite{fleming2024medalign} pairs clinician-authored instructions with expert responses grounded in longitudinal EHRs; EHRSHOT \cite{wornow2023ehrshot} evaluates few-shot prediction from structured patient data; and TIMER \cite{cui2025timer} targets temporal reasoning over longitudinal clinical records. At a broader level, MedHELM \cite{bedi2025medhelm} provides a clinician-validated framework for organizing evaluation across heterogeneous medical tasks. Collectively, these resources assess important capabilities in instruction following, temporal reasoning, and patient-specific prediction. A shared boundary, however, is that relevant patient information is typically supplied, preprocessed, or retrieved outside the evaluated decision loop. They therefore do not directly evaluate whether an agent can determine what evidence is needed, retrieve that evidence across heterogeneous EHR sources, and assess whether the available record is sufficient to support a defensible conclusion.

\subsection{Interactive Clinical Simulation and Prospective Decision-Making}

Interactive clinical environments evaluate sequential information acquisition and decision-making. AgentClinic \cite{schmidgall2024agentclinic} simulates clinical encounters in which agents interact with patients, request information, use tools, and formulate diagnoses. The Clinical Environment Simulator \cite{luo2026clinicalenvsim} extends this paradigm to digital hospitals by modeling patient-state transitions alongside operational constraints such as bed availability, staff workload, and equipment status. These environments expose failures not captured by static question answering, including inefficient information gathering, inappropriate test selection, and errors arising from sequential interactions. However, their patient states or interaction dynamics are wholly or partly simulated, and they primarily evaluate prospective diagnosis and management.

Motivated by the view that clinical AI should be evaluated within evolving real-world workflows rather than as a static model in isolation \cite{rosenthal2025rethinking}, recent benchmarks construct interactive evaluation from real clinical data. MIMIC-CDM \cite{hager2024clinicaldecision} derives sequential decision-making cases from MIMIC-IV cases, requiring models to request physical examinations, laboratory tests, and imaging before producing diagnoses and treatment plans. Its original evaluation, however, was restricted to open-access models from the Llama 2 generation, leaving the performance of newer frontier agents in this environment unresolved. MIRA \cite{ferber2026mira} constructs prospective-style encounters from MIMIC-IV-derived cases, allowing agents to elicit information, order tests, and develop diagnostic and treatment plans. ClinEnv \cite{lu2026clinenv} converts real inpatient admissions into sequential decision stages and evaluates both clinical decisions and information-acquisition behavior. These benchmarks substantially advance the evaluation of longitudinal, consequential clinical decision-making, but their primary unit of evaluation remains a prospective decision or management trajectory rather than what can be established retrospectively from an existing longitudinal record.

\subsection{Agentic Retrieval and Execution over EHR Workflow}

Research on agentic EHR interaction has progressed from executable question answering to long-horizon clinical workflows. EHRSQL \cite{lee2022ehrsql} pairs clinician-derived questions with executable SQL queries over structured EHR databases and includes questions that cannot be answered under the available schema. EHRAgent \cite{shi-etal-2024-ehragent} extends this setting by iteratively generating and executing code for multi-table EHR reasoning. MedAgentBench \cite{jiang2025medagentbench} evaluates physician-authored cases in a FHIR-compliant virtual EHR, while FHIR-AgentBench \cite{Lee2025FHIRAgentBenchBL} evaluates retrieval, interaction, and reasoning strategies for question answering over interoperable FHIR resources. EHR-Complex \cite{qiao2026ehrcomplexbenchmarkingmedicalagents} further scales executable analysis over MIMIC-IV to patient- and population-level cases requiring SQL or Python. 

More recent benchmarks evaluate longer and more composite clinical workflows. PhysicianBench \cite{liu2026physicianbench} requires agents to retrieve information across encounters, reason over heterogeneous records, perform clinical actions, and produce documentation, with performance assessed through execution-grounded checkpoints. LongMedBench \cite{chen2026longmedbench} evaluates fact retrieval, temporal reasoning, and clinical decision-making across repeated admissions and extended patient histories. Collectively, these benchmarks advance the evaluation of EHR retrieval, executable reasoning, case completion, and long-horizon workflow performance, with evaluation centered primarily on answer correctness, execution success, and completion of predefined clinical tasks.

\subsection{Evidence Grounding, Policy Adherence, and Abstention in Clinical AI}

Complementing benchmarks focused on interaction and case completion, another line of work evaluates the reliability of clinical model outputs beyond aggregate answer accuracy. VeriFact-BHC \cite{chung2026verifact} and FactEHR \cite{munnangi2024factehr} assess the factual consistency of clinical text against source records, while ArchEHR-QA \cite{soni2026archehrqa} evaluates evidence-grounded question answering over clinical notes. Related work also shows that performance on medical multiple-choice benchmarks can decline substantially when familiar answer options are replaced with a none-of-the-above choice, suggesting that high accuracy may partly reflect response-format and answer-pattern cues rather than robust evidence-based reasoning \cite{bedi2025fidelity}. These benchmarks highlight the importance of determining whether generated claims are supported by patient records and whether model conclusions are grounded in the available evidence.

Reliability also depends on whether an agent applies the appropriate clinical standard and follows case-specific investigation procedures. MIMIC-CDM \cite{hager2024clinicaldecision} evaluates guideline concordance using encoded criteria for recommended testing and treatment rather than requiring models to retrieve, cite, or interpret the guidelines. Prior work has also incorporated guideline knowledge into longitudinal EHR decision support \cite{li2025clicare} and evaluated intermediate workflow and information-acquisition behavior \cite{liu2026physicianbench,lu2026clinenv}. These studies motivate evaluation of whether agents identify and apply relevant standards, interpret thresholds, timing rules, and exceptions correctly, and follow required investigation steps without relying on invalid evidentiary shortcuts \cite{bedi2025fidelity, natmed2026robustness}.

Prior work has further examined when clinical systems should abstain from a definitive answer. EHRSQL \cite{lee2022ehrsql,lee-etal-2024-overview} evaluates whether questions can be answered under the available database schema, MedAbstain \cite{machcha2026medabstain} studies uncertainty-sensitive abstention in medical question answering, and ClinDet-Bench \cite{watanabe2026clindet} assesses whether incomplete clinical descriptions contain sufficient information to apply a clinical criterion. Together, these studies illustrate different sources of indeterminacy, including missing information and uncertainty that remains after the available evidence has been considered.

\subsection{The CliniCARE-Bench Distinction}

Existing clinical benchmarks have evaluated medical knowledge \cite{singhal2023llm,nori2023gpt4medical}, temporal reasoning \cite{cui2025timer}, factuality \cite{chung2026verifact,munnangi2024factehr}, abstention \cite{lee2022ehrsql,machcha2026medabstain,watanabe2026clindet}, and agentic EHR interaction \cite{jiang2025medagentbench,Lee2025FHIRAgentBenchBL,qiao2026ehrcomplexbenchmarkingmedicalagents}, but these capabilities are typically assessed in separate settings. CliniCARE-Bench brings them together in an end-to-end evaluation of retrospective clinical audit over real-patient-derived longitudinal EHR data \cite{physionet-mimic-iv}---to our knowledge the most comprehensive combination of these capabilities in a single clinical agent benchmark to date (Table~\ref{tab:benchmark_positioning}). Given a concise patient-level audit question, the agent must independently plan and execute retrieval across structured records and free-text notes through governed, auditable interfaces; identify and apply the governing clinical policy or adjudication criteria; and produce a cited verdict supported by a complete and traceable evidence trail. 

CliniCARE-Bench is designed as a deployment-oriented first exam for selective autonomy rather than solely as a frontier stress test \cite{sun2026agents}. It focuses on representative, evidence-intensive workflows that recur in clinical quality, protocol, and documentation review. This distinction is also motivated by the optimization paradox observed in
clinical multi-agent systems. Using the MIMIC-CDM environment of Hager
et al., Bedi et al.~\cite{bedi2025optimizationparadox} found that improvements in
component-level measures did not reliably predict end-to-end system
accuracy, including across diagnostic outcomes, process adherence, and
cost-related measures. CliniCARE-Bench therefore reports these dimensions
separately and evaluates the complete investigation-and-adjudication
trajectory rather than treating any component metric as a proxy for
clinical correctness. It assesses whether an agent can reliably resolve cases within its capabilities while distinguishing cases that lack necessary evidence from those that remain medically ambiguous after review. Accordingly, performance extends beyond final-verdict accuracy to the adequacy of evidence retrieval, faithfulness of cited claims, completeness of decisive findings, adherence to applicable rules and required investigation procedures, and appropriate escalation. This design is intended to inform human-supervised deployment by measuring both the share of work an agent can resolve autonomously and whether those decisions are sufficiently transparent, reproducible, and auditable for operational use. 

More broadly, the need for such auditable agents is not specific to medicine. Recent work argues for formalizing LLM agent security in terms of explicit, verifiable properties, such as authorized objectives, action alignment, source authorization, and data isolation, rather than treating agent trustworthiness as implicit \cite{siu2026llmagentsecurity}. CliniCARE-Bench provides a clinical instantiation of this broader agenda by combining governed, logged tool access with process-aware evaluation of evidence grounding, policy adherence, and reproducibility. This framework provides a general recipe for auditable agent environments that transfers to other high-stakes domains, such as finance, cybersecurity, and law, where agent decisions must be transparent and verifiable.

\benchmarkname{} is designed to complement existing efforts. \benchmarkname{} will be integrated into the MedHELM framework \cite{bedi2025medhelm} so that its process-aware, grounding- and abstention-scored cases can be evaluated alongside benchmarks such as PhysicianBench \cite{liu2026physicianbench} and HealthAdminBench \cite{bedi2026healthadminbench} within a shared, openly accessible evaluation ecosystem.

\begin{table}[ht]
\centering
\caption{\textbf{Positioning of CliniCARE-Bench against representative clinical-AI benchmarks.} We compare evaluation frameworks across their technical substrates and core evaluation capabilities. Checkmarks (\checkmark) denote full support, tildes ($\sim$) denote partial support, and em-dashes (---) indicate absence.}
\label{tab:benchmark_positioning}
\small
\setlength{\tabcolsep}{5pt}
\resizebox{\textwidth}{!}{%
\begin{tabular}{llcccc}
\toprule
\textbf{Benchmark} & \textbf{Substrate} & \textbf{\begin{tabular}[b]{c}Agentic\\Env.\end{tabular}} & \textbf{\begin{tabular}[b]{c}Real\\EHR\end{tabular}} & \textbf{\begin{tabular}[b]{c}Grounding\\Scored\end{tabular}} & \textbf{\begin{tabular}[b]{c}Calibrated\\Abstention\end{tabular}} \\
\midrule
AgentClinic \cite{schmidgall2024agentclinic} & Simulated patients & \checkmark & --- & --- & --- \\
HealthBench Professional \cite{soskinhicks2026healthbenchpro} & Model--clinician chat cases & --- & --- & --- & --- \\
HealthAdminBench \cite{bedi2026healthadminbench} & Admin / claims & $\sim$ & --- & --- & --- \\
MedHELM \cite{bedi2025medhelm} & Mixed EHR cases & $\sim$ & \checkmark & $\sim$ & --- \\
VeriFact-BHC \cite{chung2026verifact} & MIMIC-III notes & --- & \checkmark & \checkmark & --- \\
FactEHR \cite{munnangi2024factehr} & Multi-source notes & --- & \checkmark & \checkmark & --- \\
ArchEHR-QA \cite{soni2026archehrqa} & MIMIC-III/IV note excerpts & --- & \checkmark & \checkmark & --- \\
TIMER \cite{cui2025timer} & Longitudinal notes & --- & \checkmark & $\sim$ & --- \\
CliCARE \cite{li2025clicare} & Cancer EHR + guidelines & --- & \checkmark & --- & --- \\
MedAlign \cite{fleming2024medalign} & Stanford EHR & --- & \checkmark & $\sim$ & --- \\
EHRSHOT \cite{wornow2023ehrshot} & Stanford EHR (structured) & --- & \checkmark & --- & --- \\
EHRSQL \cite{lee-etal-2024-overview} & MIMIC-IV (text-to-SQL) & --- & \checkmark & --- & $\sim$ \\
MedAbstain \cite{machcha2026medabstain} & Medical MCQ & --- & --- & --- & $\sim$ \\
ClinDet-Bench \cite{watanabe2026clindet} & Clinical scoring cases & --- & --- & --- & $\sim$ \\
EHRAgent \cite{shi-etal-2024-ehragent} & MIMIC-III / eICU (tabular) & \checkmark & \checkmark & --- & --- \\
EHR-Complex \cite{qiao2026ehrcomplexbenchmarkingmedicalagents} & MIMIC-IV (SQL/code) & \checkmark & \checkmark & --- & --- \\
MedAgentBench \cite{jiang2025medagentbench} & Virtual FHIR EHR & \checkmark & $\sim$ & --- & --- \\
FHIR-AgentBench \cite{Lee2025FHIRAgentBenchBL} & MIMIC-IV-FHIR & \checkmark & \checkmark & --- & --- \\
MIRA \cite{ferber2026mira}       & Sim.\ EHR (MIMIC-IV)    & \checkmark & $\sim$ & --- & --- \\
ClinEnv \cite{lu2026clinenv} & Real inpatient admissions & \checkmark & \checkmark & --- & --- \\
PhysicianBench \cite{liu2026physicianbench} & Real-world EHR workflows & \checkmark & \checkmark & --- & --- \\
LongMedBench \cite{chen2026longmedbench} & Longitudinal MIMIC-IV & $\sim$ & \checkmark & --- & --- \\
\midrule
\textbf{CliniCARE-Bench (Ours)} & \textbf{MIMIC-IV (all modules)} & \textbf{\checkmark} & \textbf{\checkmark} & \textbf{\checkmark} & \textbf{\checkmark} \\
\bottomrule
\end{tabular}%
}
\begin{flushleft}
\footnotesize \textit{Note:} "Agentic Env." implies the model autonomously plans multi-step retrieval and executes tool calls in a closed loop. "Grounding Scored" indicates that model assertions are directly evaluated against granular, cited record spans. "Calibrated Abstention" denotes that the benchmark explicitly scores and rewards principled abstention, distinguishing insufficient evidence (Indeterminate: Lack of Data) from genuine medical ambiguity (Indeterminate: Medically Ambiguous), when chart evidence cannot support a definitive verdict.
\end{flushleft}
\end{table}
\section{The Benchmark Environment}
\subsection{Data Substrate}
\benchmarkname{} is constructed on top of three datasets from the MIMIC-IV database: the core MIMIC-IV v3.1, MIMIC-IV-Note v2.2, and MIMIC-IV-ED v2.2 \cite{physionet-mimic-iv, physionet-mimic-iv-note, physionet-mimic-iv-ed}. Together, these datasets provide retrospective, de-identified records from the inpatient, intensive-care, and emergency departments of a single major academic medical center. They include both structured clinical events and unstructured discharge summaries and radiology reports, organized into four modules comprising 41 tables:

\begin{itemize}
    \item \textbf{The \texttt{hosp} module (22 tables):} Contains hospital-wide encounter data, including admissions, transfers, coded diagnoses and procedures (ICD-9/10), medication ordering and administration (prescriptions/eMAR), laboratory results, microbiology, and code dictionaries.
    \item \textbf{The \texttt{icu} module (9 tables):} Captures high-frequency granular data from intensive care units, including chart events, fluid input/output balances, and procedural records.
    \item \textbf{The \texttt{ed} module (6 tables):} Preserves emergency department timelines, tracking triage vitals, initial acuity levels, and medication reconciliation upon arrival.
    \item \textbf{The \texttt{note} module (4 tables):} Provides the full unstructured text of discharge summaries and radiology reports alongside structured note-level metadata (e.g. author, exam name, CPT code).
\end{itemize}

In total, the database comprises 364,627 unique patients, 546,028 hospital admissions, 94,458 ICU stays, approximately 331,000 discharge summaries, 2.3 million radiology reports, and around 900 million structured rows. 

\subsection{Agentic Tool Environment}

While MIMIC-IV provides the underlying data substrate, \benchmarkname{} does not expose its physical database schema as the primary interface. Instead, agents interact with patient records through a governed tool layer organized around clinically meaningful entities and retrieval operations. This design reflects a common production access pattern in which clinical AI systems interact with EHR data through application-layer interfaces, such as Fast Healthcare Interoperability Resources (FHIR)-based interfaces or vendor-specific APIs, rather than through unrestricted access to raw database tables. Our environment is not intended to reproduce a specific interoperability standard. Instead, it captures three operational properties central to evaluation.

First, access through the primary clinical tool surface is scoped to one patient at a time. Rather than receiving a preassembled record, the agent retrieves targeted portions of the patient’s chart through bounded, parameterized tools. These tools organize access around clinically meaningful record categories, including demographics and encounters, notes, laboratory results, vital signs, medications, microbiology, procedures, imaging, and orders, rather than around MIMIC-IV’s physical table structure. Second, the environment preserves the heterogeneity of the source record. Data are neither pre-joined nor pre-summarized, requiring the agent to reconcile structured events with free-text documentation and reconstruct the relevant clinical timeline. Third, every tool invocation is logged and replayable, enabling the benchmark to inspect which evidence the agent retrieved and whether its final report is supported by that evidence.

Together, these properties make the agent's path through the record, not just its final answer, part of what the benchmark measures. The tool surface comprises the following sets of tools:

\begin{itemize}
    \item \textbf{Clinical EHR Tools:} Parameterized, clinician-verified tools which support common record-retrieval operations across the MIMIC database, including patient and encounter discovery, note retrieval and search, and longitudinal access to laboratory results, vital signs, medication administration, microbiology, procedures, imaging, and provider orders. These tools constitute the primary interface to the clinical record and organize access around clinically meaningful entities rather than the underlying database schema.

    \item \textbf{SQL Fallback:} A capped, read-only SQL interface retained to address coverage gaps in the higher-level clinical tools. Its use is logged separately so that the benchmark can quantify when an agent relies on the underlying database schema rather than the intended clinical interface.

    \item \textbf{Shell Execution:} A sandboxed execution environment with preconfigured analytical libraries, including \texttt{pandas}, \texttt{numpy}, and \texttt{scipy}. Clinical adjudication frequently requires derived quantities, like a computed severity score, an event interval, or a trended value, which are not stored in the record. The sandbox allows the agent to compute these quantities explicitly and reproducibly, with commands, outputs, and generated artifacts retained as part of the run trace for inspection. 
        
    \item \textbf{Policy-Document Tools:} A lightweight file-access interface for listing, searching, and reading documents in the given policy corpus. Retrieved passages retain document and line-level provenance, enabling policy-dependent claims to be traced to the source text. The policy corpus and grounding procedure are described in Section~\ref{sec:policy}.

\end{itemize}

\subsection{Policy and Guideline Grounding}
\label{sec:policy}

Many clinical audit and decision-support cases require context that extends well beyond the patient record. While EHR artifacts capture what occurred during an encounter, evaluating and contextualizing that data requires an external standard. The evaluation may involve determining whether an action was appropriate, timely, or compliant; establishing the diagnostic or laboratory thresholds a determination hinges on; or applying a governing classification, coding, or quality-measure definition. Many \benchmarkname{} cases are therefore not purely factual questions about the EHR, but also require patient-level evidence to be interpreted under an applicable clinical, regulatory, or operational standard.

We refer to this standard as a \textbf{governing policy}. A governing policy may be an institutional protocol, a professional-society guideline, a regulatory or payer quality measure, or another formally adopted clinical standard. These policies define the thresholds, timing requirements, exception criteria, and decision rules against which patient-level evidence must be interpreted. For example, a sepsis audit may depend on the inclusion and timing criteria specified by CMS SEP-1 \cite{SevereSepsisSeptic}, and an anticoagulation-reversal audit may require comparison against the American College of Cardiology's expert-consensus treatment recommendations \cite{tomaselli2020ACCExpert2020}. Without access to the applicable policy, an agent may retrieve the patient data correctly yet still reach the wrong verdict, applying generic, outdated, or contextually inappropriate reasoning. The governing policy is what supplies the decision criteria the evidence is judged against. Therefore, policy grounding is not an auxiliary retrieval step but part of the adjudication itself, allowing the benchmark to distinguish a policy-supported verdict from one based on unsupported assumptions or coincidentally correct reasoning.

To evaluate this capability, \benchmarkname{} includes a policy-grounding component that tests whether an agent can identify, retrieve, cite, and correctly apply authoritative documentary evidence alongside the clinical record. Agents are given a fixed corpus of policy documents curated with input from the Clinical Board. Since the corpus is shared across all cases rather than filtered into case-specific subsets, the agent must itself determine which document governs the question and locate the relevant provisions. For evaluation, each scenario specifies its own governing document or document set. The documents are converted from PDF to Markdown and exposed through the policy-document tools described earlier, which list, search, and read document text and section metadata. Retrieved passages retain document- and line-level provenance, so each policy-dependent claim can be traced back to specific source text.

This makes the incorporation of governing standards into the agent's adjudication both reproducible and clinician-inspectable, and it mirrors a practical requirement of clinical AI deployment, in which an agent must not only retrieve relevant patient information but also identify which policy applies and make its interpretation auditable. Policy grounding therefore complements EHR grounding, connecting observed clinical events to the standards against which those events are judged.
\section{Scenario and Case Construction}

\begin{figure}[t]
\centering
\includegraphics[width=\linewidth]{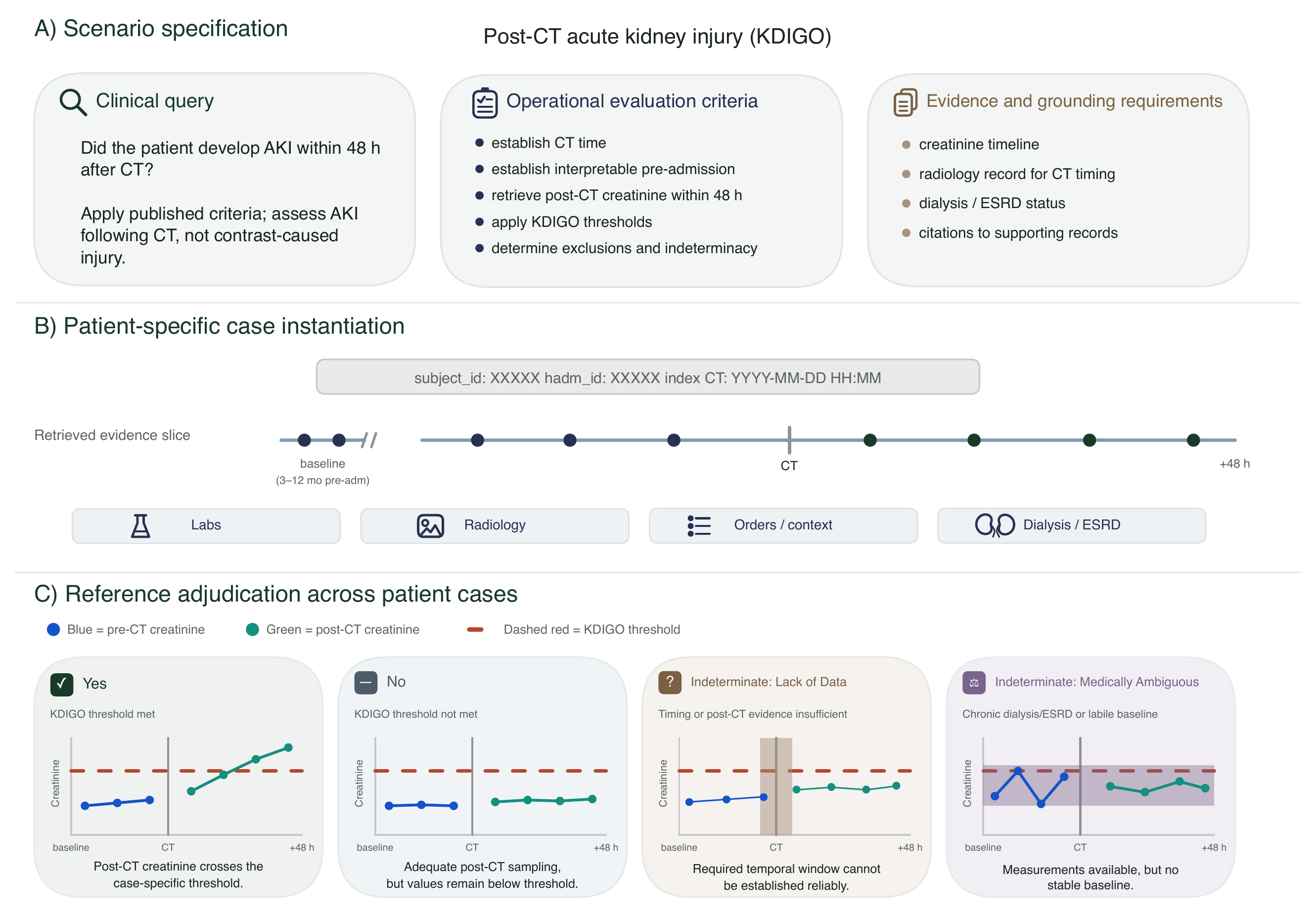}
\caption{\textbf{From scenario specification to patient-specific adjudication.}
\textbf{(A)} The post-CT acute kidney injury (AKI) scenario defines its clinical query, operational evaluation criteria, and evidence and grounding requirements.
\textbf{(B)} These are instantiated for a specific case (shown schematically) from the encounter, CT event, and retrieved EHR evidence.
\textbf{(C)} The same KDIGO serum-creatinine criteria yield one of four reference verdicts---\textit{Yes}, \textit{No}, \textit{Indeterminate: Lack of Data}, or \textit{Indeterminate: Medically Ambiguous}. Blue and green mark pre- and post-CT creatinine; the dashed red line is the case-specific KDIGO threshold ($\geq$0.3\,mg/dL or $\geq$1.5$\times$ baseline).}
\label{fig:scenario-anatomy}
\end{figure}

The design of \benchmarkname{} begins with the construction of \emph{clinical scenarios}. A scenario translates a clinically meaningful audit or chart-review question into a standardized evaluation specification: it defines the question presented to the agent, the
applicable clinical or policy criteria, the evidence required for adjudication, the conditions associated with each verdict, and the reasoning procedures and shortcuts that should be rewarded or penalized.

Each scenario centers on a clinical proposition that could in principle be affirmed or rejected, but that the benchmark adjudicates with one of four verdicts: \textit{Yes}, \textit{No}, \textit{Indeterminate: Lack of Data}, or \textit{Indeterminate: Medically Ambiguous}. The first two indicate that the available record supports a definitive adjudication under the scenario criteria. \textit{Indeterminate: Lack of Data} applies when evidence required for adjudication is absent or cannot be reliably
established from the record. \textit{Indeterminate: Medically Ambiguous}
applies when the relevant evidence is available but does not support a
unique, clinically defensible interpretation. This shared output space
makes abstention part of the evaluation standard.

Since it is standardized, a scenario specification applies consistently across heterogeneous patient records. As such, each scenario is instantiated over a cohort of eligible MIMIC-IV records to create \emph{patient-specific
cases}. \benchmarkname{} contains 30 cases per
scenario, for a total of 750 cases. The case-level
reference verdicts are produced through the clinician-calibrated labeling procedure described in Section~\ref{sec:clinical-calibration}.

The remainder of this section
describes the anatomy of a scenario, presents representative scenarios,
explains how scenarios are instantiated as patient cases, and details the
clinical review and label-calibration procedure. Terminology and units of
evaluation are summarized in Appendix~\ref{app:terminology}.

\subsection{Scenario Development}
\begin{figure}[ht]
\centering
\includegraphics[width=\linewidth]{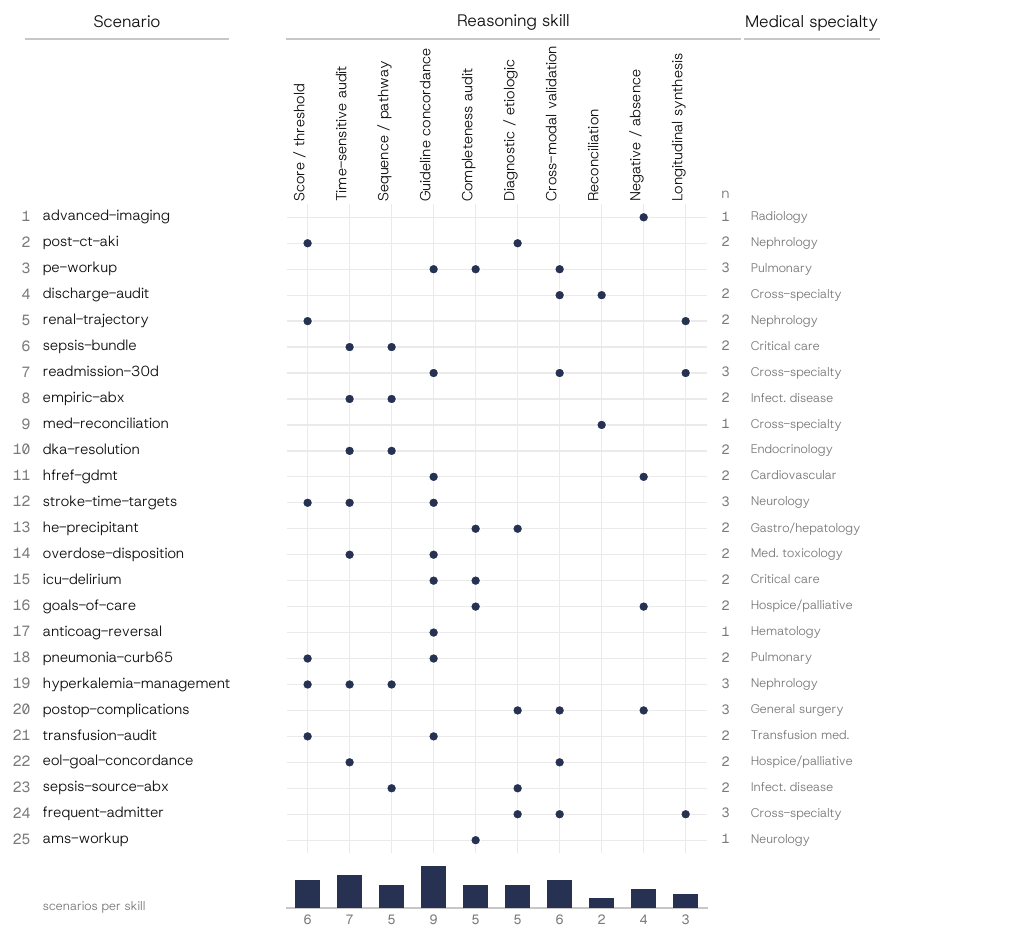}
\caption{\textbf{Reasoning-skill and specialty coverage across the 25 CliniCARE-Bench scenarios.} Each row is one scenario, and a filled dot marks a
reasoning skill that the scenario exercises. Most scenarios probe two or three
skills at once ($n$ column = skills per scenario), so the benchmark
targets compound clinical reasoning rather than isolated skills. The bars below
each column count the scenarios exercising that skill. Decision-tree / guideline
concordance ($9$) and time-sensitive protocol audit ($7$) are the most widely
sampled, whereas discrepancy / reconciliation ($2$) and longitudinal synthesis
($3$) are rarer, higher-difficulty probes. The final column gives each
scenario's primary ABMS specialty. Because a given reasoning skill recurs across
many specialties, the skill and specialty axes are near-orthogonal, so the same
capability is not tied to one organ system or workflow.}
\label{fig:coverage}
\end{figure}

Our suite comprises \emph{25 clinician-developed and validated scenarios} spanning recurring clinical audit and review workflows. The suite was designed to
provide complementary coverage across both reasoning capabilities and medical specialties. Its reasoning demands span ten capabilities, enumerated in Figure~\ref{fig:coverage}, from clinical-score and threshold computation and time-sensitive protocol auditing to cross-modal validation between documentation and structured data, negative and absence reasoning, and longitudinal synthesis across encounters.

Rather than defining an ad hoc domain grouping, we label each scenario with a
single primary clinical specialty drawn from the American Board of Medical
Specialties (ABMS) taxonomy, which defines 40 specialties and 87 subspecialties
(127 certificate categories in
total).\footnote{\url{https://www.abms.org/member-boards/specialty-subspecialty-certificates/}}
The 25 scenarios span 14 specialties: nephrology, infectious disease, critical
care medicine, pulmonary disease, cardiovascular disease, neurology (including
vascular neurology), endocrinology, gastroenterology and hepatology, hematology,
transfusion medicine, medical toxicology, hospice and palliative medicine,
general surgery, and radiology. Four scenarios (discharge-audit, readmission-30d,
med-reconciliation, and frequent-admitter) are quality, documentation, and
care-coordination workflows not owned by a single specialty. We group these as
\emph{cross-specialty operations}. Specialty is a single-label axis, where each scenario has exactly one primary specialty, in contrast to the reasoning-skill axis, which
is non-exclusive, since a scenario may exercise several reasoning capabilities.

Figure~\ref{fig:coverage} gives the full coverage matrix along both
axes. Because specialty is single-label while reasoning skill is non-exclusive,
the two views are complementary: the specialty axis conveys clinical breadth,
whereas the skill axis shows that the same reasoning capability recurs across
specialties rather than being tied to one organ system or workflow. Full specifications for four representative scenarios appear in Appendix~\ref{app:cases}, and
Table~\ref{tab:secondary_case_selection} highlights a representative subset.

\subsection{Anatomy of a Scenario}
\label{sec:scenario-anatomy}

Each scenario is defined by the following components:

\begin{itemize}

\item \textbf{Clinical Query:}
A concise question expressed in the natural language familiar to the intended clinical user. The query specifies the clinical objective while leaving the detailed retrieval and analysis strategy to the agent.  At the scenario level, the
query contains placeholders such as \texttt{subject\_id},
\texttt{hadm\_id}, or \texttt{stay\_id}, which are populated when the
scenario is instantiated as a patient-specific case.

\item \textbf{Operational Evaluation Criteria:} A detailed specification of the decision rules that map the available
patient record to one of four verdicts: \textit{Yes}, \textit{No},
\textit{Indeterminate: Lack of Data}, or
\textit{Indeterminate: Medically Ambiguous}. These criteria define the
applicable thresholds, temporal windows, exclusions, exception rules, and
conditions under which the record is insufficient or clinically ambiguous.

\item \textbf{Evidence and Grounding Requirements:}
A specification of the evidence needed to support adjudication, including
the relevant record sources, clinical variables, temporal windows, and the
governing policy documents. These requirements define
the decisive findings that the agent must retrieve and cite, or explicitly
establish as unavailable, in order to justify its verdict. Three of these
requirements are recorded as explicit per-scenario lists, and serve as the
reference sets for the grounding metrics of Sections~\ref{sec:patient-grounding}
and~\ref{sec:policy-grounding-metrics}: the \emph{required record sources},
three to nine MIMIC-IV tables per scenario over a vocabulary of 23 tables; the \emph{required
findings}, four to eleven distinct content dimensions per scenario; and the
\emph{governing documents}, one to four per scenario drawn from
the policy corpus of Section~\ref{sec:policy}.

\end{itemize}

The clinical query is presented to the agent, while the operational
criteria are applied through the labeling procedure to produce case-level reference verdicts. The evidence and grounding requirements define the information needed to support adjudication and enable evaluation beyond verdict correctness. Together, these components define a standardized four-way clinical adjudication problem with an explicitly specified evidentiary basis.

\subsection{Representative Scenarios}
To make the scenario structure concrete, we summarize one representative scenario below. Figure~\ref{fig:scenario-anatomy} illustrates how the scenario is instantiated and adjudicated for an individual patient case.  The complete operational specification for this scenario, clinical query, four-way adjudication criteria, required evidence, grounding requirements, and adversarial controls, together with three additional fully annotated anchor scenarios, is provided in Appendix~\ref{app:cases}. These per-case specifications are a central contribution of this work.

\paragraph{Post-CT Acute Kidney Injury.}
This scenario asks whether an ICU patient developed acute kidney injury
within 48 hours after a CT scan, applying the published KDIGO criteria.
Baseline creatinine is set by a three-tier hierarchy: the lowest value
3--12\,months before admission, else the admission value, else an MDRD
estimate at eGFR 75\,mL/min/1.73\,m$^2$ where CKD is documented.

\begin{itemize}
    \item \textbf{Clinical Query:} ``For ICU patient \texttt{<subject\_id, hadm\_id>} who underwent a CT this admission, did the patient develop AKI within 48\,h after the CT? [\ldots] Apply the KDIGO serum-creatinine criteria and frame the result as AKI following CT, not as contrast-caused injury. Where serum creatinine does not reflect the patient's own kidney function, say the question does not resolve rather than applying the thresholds.'' The elided passage states the baseline hierarchy above; the full prompt appears in Appendix~\ref{app:cases}.
    \item \textbf{Adjudication Criteria:} A \emph{qualifying rise} is any post-CT creatinine at least $0.3$\,mg/dL above baseline or at least $1.5\times$ baseline; every value in the window is searched, not only the first.
    \begin{itemize}
        \item \textit{Yes:} a baseline is set, at least one creatinine falls within 48\,h post-CT, there is a qualifying rise, and the patient is not on chronic dialysis and has no ESRD.
        \item \textit{No:} a baseline is set, at least two creatinines fall within 48\,h post-CT, there is no qualifying rise, and the patient is not on chronic renal replacement therapy and has no ESRD or CKD stage $\geq 4$.
        \item \textit{Indeterminate: Lack of Data:} no tier of the baseline hierarchy can be set; or no creatinine falls within 48\,h post-CT, or fewer than two, leaving the peak unconfirmed; or CT timing cannot be resolved (order versus actual scan time, or no unique first qualifying CT); or no CT can be identified in the admission, so the premise is unmet and the case is out of scope.
        \item \textit{Indeterminate: Medically Ambiguous:} chronic dialysis or ESRD, so the KDIGO acute thresholds do not apply; or a labile pre-admission baseline, so no clean reference can be set and a $+0.3$\,mg/dL rise cannot be confidently attributed.
    \end{itemize}
    \item \textbf{Required Evidence:} the creatinine timeline from the baseline source through 48\,h post-CT, showing every value; ESRD and dialysis status; and the CT \emph{scan} time distinguished from the order time, confirmed against the radiology report where possible.
    \item \textbf{Required Grounding:} The specific MIMIC-IV record or record span from which the decisive finding was derived.
\end{itemize}

\subsection{From Scenarios to Patient Cases}

To instantiate a scenario, we first construct a scenario-specific
candidate cohort from MIMIC-IV. Broad eligibility criteria identify
records containing the encounters, index events, and evidence sources
relevant to the clinical question. Because a case may be anchored to a
particular admission, ICU stay, procedure, or other clinical event, case
eligibility is defined at the level appropriate to the scenario.

We then apply scenario-specific heuristics to assign each eligible record
to a provisional sampling stratum. These heuristics are used only for
cohort construction and do not determine the final reference verdict.
We sample 30 patients per scenario using stratification across
\textit{Yes}, \textit{No}, and \textit{Indeterminate} candidate strata,
with the Indeterminate stratum including cases provisionally associated
with either lack of data or medical ambiguity. This intentional balancing
prevents aggregate performance from being dominated by straightforward
positive or negative cases and increases coverage of scenario-specific
challenges, including missing evidence streams, conflicting
documentation, ambiguous temporal attribution, and potentially
misleading administrative codes. As such, the resulting case distribution is designed for evaluation and should not be interpreted as an estimate of
the clinical prevalence of these outcomes.

Case-level reference verdicts are then produced through a model-assisted
labeling procedure. An ensemble of three frontier model harness configurations from separate families, namely \texttt{Claude Code} Opus~4.8, \texttt{Codex} GPT-5.5, and \texttt{Gemini CLI} Gemini~3.1~Pro, independently
evaluate each case with access to the complete scenario specification,
including the clinical query, operational evaluation criteria, and
evidence and grounding requirements. Their outputs are aggregated to
produce the reference verdict. Across the 750 cases the agents reported unanimously on 556, split 2-vs-1 on 183, and produced three distinct verdicts on 11, with only the last set being escalated to clinicians for direct adjudication. This process is  validated through the Clinical Board procedure described in 
Section~\ref{sec:clinical-calibration}.

\subsection{Clinical Review and Label Calibration}
\label{sec:clinical-calibration}
Clinician review through our Clinical Board was used to validate each step of the pipeline. All Clinical Board reviewers are credentialed PhysioNet users who completed
the required training and executed the PhysioNet Credentialed Health Data Use Agreement before reviewing any MIMIC-derived evidence. This review
proceeded in two stages: first reviewing the construction of the 25 scenarios, then
validating a sample of the model-generated reference verdicts.

\emph{Scenario review:} two clinicians independently vetted every scenario, confirming the query reflects a real chart-review workflow, estimating manual adjudication time, and revising the four-way criteria (thresholds, windows, exclusions, ambiguity conditions, and evidence requirements). Each decision rule was tagged by provenance (i.e., named guideline or policy, established clinical knowledge, or benchmark-specific operational choice), making explicit which parts of the adjudication are externally governed. These edits were applied before the ensemble was run. 

\emph{Verdict calibration:} After verdicts were generated, a sample of them was shown to clinicians to validate the generation process. In this review, clinicians were given the scenario specification and the three ensemble reports in randomized order as Alpha, Beta, and Gamma to remove model-identity bias, then asked to return their own four-way verdict with a confidence and rationale. We note that this is not truly independent, as the reviewer uses the model-surfaced evidence to make their judgment instead of interfacing with the patient record themselves. This design is intentional: we rely on the models for the SQL and data-extraction work while reserving for the Clinical Board the medical judgment of interpreting that evidence. The reviewer also rated each report as \emph{sound}, \emph{minor error}, or \emph{major error}, tagging errors by category (missed evidence, misapplied criteria, temporal or attribution error, unsupported inference, or incorrect abstention). 

We ran verdict calibration at two scales: a 75-case sample (10\%; three cases per scenario) with one reviewer each for verdict-level calibration, and a 25-scenario subset (one case each) with two independent reviewers, with no reviewer seeing the same case twice, to measure reproducibility. On the 75-case sample, the reference verdict matched the clinician verdict on 91\% of cases with a defined label (64/70; Cohen's $\kappa = 0.87$). The five no-majority cases were escalated by design. Agreement was perfect when the ensemble was unanimous (46/46), and fell to 75\% (18/24) on 2-vs-1 cases. Every majority-case disagreement resolved to the dissenting model rather than an outside label, with expert judgment staying within the ensemble's range.

The double-reviewed subset places these figures against the reproducibility of expert judgment itself. The two reviewers agreed with each other on 80\% of verdicts ($\kappa = 0.69$), while the reference agreed more closely with each of them individually ($\kappa = 0.87$ and $0.81$): the model-assisted label sits closer to each clinician than the two clinicians sit to one another. Agreement again tracked ensemble consensus, and reviewer-reviewer agreement fell fastest on the non-unanimous cases. Non-unanimity therefore marks the cases where expert judgment is itself least reproducible, supporting their escalation. Reviewers rated most reports \emph{sound}: 73\% of the 225 report ratings on the 75-case sample and 81\% of the 150 on the double-reviewed subset, with the remainder split between minor and major errors. These ratings were consistent across reviewers (Gwet's AC1~$\approx0.70$; Cohen's $\kappa$ understates agreement here because \emph{sound} dominates the marginal). Among the errors reviewers did flag, misapplied criteria was the most common category (38\% and 53\% of tags) ahead of missed evidence (24\% in both), locating the ensemble's failures in applying a scenario's decision rules rather than in surfacing the underlying evidence.
\section{Evaluation Framework}
\label{sec:eval_framework}

\benchmarkname{} evaluates clinical agentic systems using both their final outputs and their observable interaction traces. Each run produces a final report, a logged sequence of tool calls, the retrieved patient and policy evidence, and any computations or intermediate artifacts generated during the investigation. These artifacts are evaluated against the case-level reference verdict and scenario-specific evaluation metadata across various complementary dimensions, including verdict correctness, abstention behavior, patient-evidence grounding, policy grounding, process adherence, and reliability and resource use.

\subsection{Units of evaluation}
\label{sec:units}

A \emph{run} is one execution of a system on one case and the unit at which metrics are scored. In our primary protocol each case is run once, giving a one-to-one correspondence between cases and runs, so per-run accuracy is computed over all 750 runs. Under $k$ repeated attempts a case contributes $k$ runs, indexed by an \emph{attempt} number and used for reliability metrics such as $\mathrm{avg}@k$ and $\mathrm{pass}\mbox{\textasciicircum}k$. A \emph{system} is the complete evaluated configuration, including the model, agent harness, prompting or scaffolding, and available tools. It is the entity that a run executes and that the benchmark ultimately characterizes.

Not every metric applies to every run. Verdict accuracy retains the entire set of runs. A \emph{report-present run} is one in which the system returns a parseable final report, and it forms the denominator for process adherence, defect-free accuracy, and the defect gap. A \emph{grounding-scoreable run} is a report-present run in which the system retrieved at least one patient record, and the four patient-grounding metrics of Section~\ref{sec:patient-grounding} are averaged over this set.

\subsection{Verdict and Abstention Metrics}
\label{sec:verdict-metrics}

Let $N$ denote the number of evaluated runs, indexed by $i \in \{1,\ldots,N\}$. Let $y_i$ denote the case-level reference verdict for run $i$, and let $\hat{y}_i$ denote the predicted verdict.

\paragraph{Four-way verdict performance.} Verdict accuracy measures exact agreement between $\hat{y}_i$ and $y_i$ across four possible verdicts: \textit{Yes}, \textit{No}, \textit{Indeterminate: Lack of Data}, or \textit{Indeterminate: Medically Ambiguous}. Verdicts are scored by exact four-class match, with no partial credit. Missing or unparseable verdicts are counted as incorrect rather than excluded from the denominator. We report four-class accuracy and macro-F1 across the four verdict classes, using Macro-F1 to give equal weight to the two less frequent, safety-critical \textit{Indeterminate} classes and the definitive \textit{Yes} and \textit{No} classes.

\paragraph{Abstention behavior.} We separately characterize whether the system defers appropriately. Let
\[
\mathcal{D}
=
\{\textit{Yes}, \textit{No}\}
\]
denote the set of definitive verdicts, and let
\[
\mathcal{A}
=
\{\textit{Indeterminate: Lack of Data},
  \textit{Indeterminate: Medically Ambiguous}\}
\]
denote the set of abstention verdicts.

Let
\[
N_{\mathcal{A}}
=
\sum_{i=1}^{N}\mathbf{1}[y_i \in \mathcal{A}]
\qquad\text{and}\qquad
N_{\mathcal{D}}
=
\sum_{i=1}^{N}\mathbf{1}[y_i \in \mathcal{D}]
\]
denote the numbers of reference-abstention and reference-definitive runs, respectively.

The \emph{over-commitment rate} is the fraction of reference-abstention runs for which the system returns a definitive verdict:
\[
\mathrm{OverCommit}
=
\frac{1}{N_{\mathcal{A}}}
\sum_{i=1}^{N}
\mathbf{1}[y_i \in \mathcal{A}]
\mathbf{1}[\hat{y}_i \in \mathcal{D}].
\]
The \emph{over-abstention rate} is the fraction of reference-definitive runs for which the system returns an abstention verdict:
\[
\mathrm{OverAbstain}
=
\frac{1}{N_{\mathcal{D}}}
\sum_{i=1}^{N}
\mathbf{1}[y_i \in \mathcal{D}]
\mathbf{1}[\hat{y}_i \in \mathcal{A}].
\]

\subsection{Patient-Evidence Grounding}
\label{sec:patient-grounding}

Patient-evidence grounding evaluates whether the system's clinical claims are supported by the patient record and whether the investigation retrieves and reports the evidence required for adjudication. We evaluate grounding along two complementary axes: \emph{citation precision}, which measures the faithfulness of the evidence cited in the report, and \emph{evidence coverage}, which measures the completeness of evidence acquisition and reporting. Each axis pairs a referential check, decidable from the trace by rule, with a substantive one that requires a semantic judgment about the report's content.

\paragraph{Citation precision.} We calculate two types of citation precision. \emph{Record-identifier precision} $P_{\mathrm{record}}$ is the fraction of record-level identifiers (e.g. patient, admission, ICU-stay, and note identifiers) cited within the final report that occur in evidence retrieved during the same run. \emph{Claim-level precision} $P_{\mathrm{claim}}$ then checks whether the evidence a claim cites substantively supports it. This distinction is necessary because a citation may refer to a real retrieved record yet still be irrelevant to, inconsistent with, or insufficient to support the associated claim.

Claim-level precision is computed by decomposing the report into atomic clinical claims, such as measured laboratory and vital-sign values, medication doses, imaging and procedure findings, temporal relationships, and the presence or absence of clinical events. Each claim is classified by an LLM against the evidence it cites as \emph{supported}, \emph{contradicted}, or \emph{unverifiable}. $P_{\mathrm{claim}}$ is the supported fraction. A claim is supported only when the cited evidence substantiates it with the correct patient and encounter attribution and a consistent temporal context; claims that the cited evidence contradicts, that cannot be verified from it, or that carry no resolvable citation are not.

\paragraph{Evidence coverage.} Similarly, coverage is assessed at two points where required evidence can be lost. \emph{Retrieval coverage} $R_{\mathrm{retrieval}}$ is the fraction of a case's required record sources that the agent queried with the correct patient and, where applicable, encounter and temporal scope. Here, required sources are the MIMIC-IV tables named in the evidence specification of the case's scenario (Section~\ref{sec:scenario-anatomy}), three to nine per scenario. A source counts as consulted when a correctly scoped query is issued, whether it returns records or a valid empty result, because an empty result may itself establish that the evidence is unavailable, a conclusion several scenarios require.

\emph{Finding coverage} $R_{\mathrm{finding}}$ is the fraction of a case's required findings that the report addresses, either by presenting the relevant evidence or by explicitly asserting its absence where that is what the case requires. Required findings are drawn from the same scenario specification, four to eleven per scenario, and are compared to the report using an LLM judge. These are, again, different failure modes: an agent may query every required source and still not establish the finding its verdict rests on.

\subsection{Policy Grounding}
\label{sec:policy-grounding-metrics}

Policy grounding is scored separately from patient-evidence grounding, and asks whether a report's use of the governing standard is verifiable: whether its policy citations resolve to real passages that support the claims they carry, and whether the report identifies the standards that actually govern the case. Every scenario carries at least one governing document, so these metrics are computed across the whole suite. Whether the agent then applies the policy's thresholds, timing requirements, exclusions, and exceptions correctly is not scored here; that is captured by verdict correctness and by the process rubric of Section~\ref{sec:process-rubric}. As in Section~\ref{sec:patient-grounding}, each axis pairs a referential check with a substantive one.

\paragraph{Citation correctness.} Agents cite policy inline as \texttt{[policy: document L\_a--L\_b]}, naming a corpus document and a line span within it. A citation \emph{resolves} when that document is present in the corpus and the span is valid, and the \emph{citation-resolution rate} is the fraction of a run's policy citations that resolve. Citations naming an absent document or an invalid span do not resolve; because a single run can emit many, we also report unresolved citations as raw counts. \emph{Citation support} then asks, for each resolved citation, whether the cited passage substantiates the particular policy-dependent claim it is attached to, rather than merely discussing a related topic. Resolution is computed by rule, while support is LLM-judged.

\paragraph{Governing-document coverage.} For run $i$, let $\mathcal{G}_i$ denote the scenario's governing-document set and $\hat{\mathcal{G}}_i$ the set of corpus documents that the report cites. \emph{Document precision} $P_{\mathrm{doc}}$ is the fraction of $\hat{\mathcal{G}}_i$ lying in $\mathcal{G}_i$, and \emph{document recall} $R_{\mathrm{doc}}$ is the fraction of $\mathcal{G}_i$ that the report cites; both are set comparisons against an expert-specified reference and require no judge. Precision asks whether the documents a report invokes govern the case at all, recall whether it found all of them, and the two can diverge sharply: a report resting on one applicable standard while omitting the others scores high precision at low recall. The two are averaged over different sets of runs, since precision is undefined for a run that cites no corpus document and that run is omitted from it, whereas recall is defined for every run and scores zero.

\subsection{Process Adherence}
\label{sec:process-rubric}

Verdict correctness and grounding do not fully establish that an agent's investigation was defensible. An agent may reach the correct verdict through an inappropriate shortcut, by omitting a verification the case turns on, or by making an unsupported inference that happens to agree with the reference verdict. Conversely, it may conduct a sound investigation and still reach the wrong verdict through a localized interpretation error. We therefore score the agent's \emph{observable investigation process} separately from its outcome. Here, observable means we use the agent's final report together with its logged tool-call trajectory; we make no attempt to inspect or score the model's private reasoning.

\paragraph{The process rubric.} Each scenario carries a \emph{process rubric}, authored by the Clinical Board and instantiated unchanged for every case of that scenario. \textsc{Must-do} criteria state the actions and checks required for a defensible adjudication. \textsc{Must-not} criteria state prohibited shortcuts and unsupported inferences, such as inferring an endpoint from an administrative code alone, treating an order time as a performed-procedure time, or asserting a causal relationship the record does not support. We call a violated \textsc{must-not} criterion a \emph{process defect}. Rubrics carry four to nine \textsc{must-do} and three to eight \textsc{must-not} criteria per scenario, each weighted $\pm1$ or $\pm2$ by the authoring clinicians.

Criteria are graded by an LLM judge against the final report and the logged trajectory together. A methodology claim in the report is credited only when the trajectory corroborates it; where the trajectory shows the agent did something other than what the report describes, the criterion is graded on what the agent actually did.

\paragraph{Process-adherence score.} For run $i$, let $\mathcal{M}_i$ denote the set of applicable criteria, each carrying a signed weight $w_m$ that is positive for \textsc{must-do} and negative for \textsc{must-not}, and let $z_{im}=1$ when the judge grades criterion $m$ as met, meaning the required action was performed or the prohibited one committed. The \emph{process-adherence score}, abbreviated \emph{process score} in the results, is

\[
S_{\mathrm{process}}^{(i)}
=
\max\left(
0,\;
\frac{
  \sum_{m \in \mathcal{M}_i} w_m z_{im}
}{
  \sum_{m \in \mathcal{M}_i \,:\, w_m > 0} w_m
}
\right),
\]

so credit accrues against the \textsc{must-do} budget while sprung \textsc{must-not} criteria subtract from it. The denominator is the sum of positive weights alone, between 7 and 16 points depending on the scenario. The floor at zero is not vacuous: in one scenario the available penalties exceed the entire positive budget. For a system we report the mean of $S_{\mathrm{process}}^{(i)}$ over report-present runs.

\paragraph{Defect-free accuracy.} A run may reach the correct verdict while springing a process defect. We therefore record for each run whether any applicable \textsc{must-not} criterion was violated,

\[
\tau_i
=
\mathbf{1}
\left[
\sum_{m \in \mathcal{M}_i \,:\, w_m < 0} z_{im} > 0
\right],
\]

and report a \emph{defect-free accuracy}, which credits a run only when its verdict is correct and $\tau_i=0$, together with the \emph{defect gap} this opens against accuracy on the same report-present denominator.

\subsection{Repeated-Run Reliability}
\label{sec:reliability}

A single-attempt score conflates whether a system can reach the correct verdict with whether it does so consistently. For a cohort evaluated over $k$ separately executed attempts, let $\hat{y}_{i,r}$ denote the verdict returned for case $i$ on attempt $r$. Because $\mathrm{pass}\mbox{\textasciicircum}k$ is undefined for a case that lacks a gradeable verdict in some attempt, the reliability metrics are computed over the $n$ cases gradeable in every attempt.

We report $\mathrm{avg}@k$, the mean verdict accuracy across all $nk$ runs, together with the standard deviation of the $k$ per-attempt accuracies, which distinguishes run-to-run noise from systematic error. We additionally report $\mathrm{pass}\mbox{\textasciicircum}k$, the proportion of cases answered correctly on every attempt:

\[
\mathrm{pass}\mbox{\textasciicircum}k
=
\frac{1}{n}
\sum_{i=1}^{n}
\prod_{r=1}^{k}
\mathbf{1}
\left[
  \hat{y}_{i,r}=y_i
\right].
\]

For comparison, $\mathrm{pass}@k$ is the proportion answered correctly on at least one attempt. The two bound reliability from either side: $\mathrm{pass}@k$ measures whether repeated attempts ever produce the correct verdict, $\mathrm{pass}\mbox{\textasciicircum}k$ whether they always do.

Neither separates a system that varies from one that is consistently wrong, so we also report \emph{verdict agreement}, the proportion of cases on which all $k$ attempts return the same verdict, correct or not. Agreement decomposes as $\mathrm{pass}\mbox{\textasciicircum}k$ plus the proportion of cases whose attempts agree on an \emph{incorrect} verdict; it is this second term that identifies a system reproducing an error rather than resolving the case.

\subsection{Metric Adjudication and Judge Reliability}
\label{sec:judge-reliability}

The framework pairs deterministic checks with LLM adjudication, and each metric above states which it uses. The division is consistent: checks that can be settled referentially, against the trace or the policy corpus, are computed by rule, while judgments about whether evidence supports a claim, whether a required finding is addressed, or whether a process criterion holds are delegated to a judge. The production judge is GPT-5.5, which is itself among the evaluated systems, so Section~\ref{sec:judge-validation} examines the resulting dependencies. Within each metric the judge model, prompt, and rubric are held fixed across evaluated systems, so a difference in score reflects the systems rather than the grader.

Each judged component receives only the evidence its judgment requires: the final report, together with the cited patient evidence, the cited policy passage, or the logged trajectory as applicable. Trajectories are rendered for judging rather than passed verbatim. Each tool result is truncated to 12{,}000 characters and the duplicate copies that some harness trajectory formats emit are dropped; where report and trajectory together still exceed roughly 500{,}000 characters, they are graded in sequential chunks and the per-criterion grades combined disjunctively, so a criterion counts as met if any chunk supports it. The report always occupies the first chunk, so criteria decided from the report alone are unaffected.

We assess the reliability of the judged metrics by regrading identical outputs and, for process and policy scoring, by repeated evaluation with a panel of judges. Section~\ref{sec:judge-validation} reports criterion-level disagreement, within-judge variation, and chance-corrected agreement statistics. Criteria whose grades prove persistently unstable are returned for expert revision rather than treated as stable measurements.

\subsection{Resource Use and Cost}
\label{sec:efficiency}

For each run we record the resources the investigation consumed: the number of agent turns, the number of typed tool calls, token volume, and monetary cost, taken from the harness where it reports one and computed from measured tokens otherwise. Token volume is summarized as \emph{work-tokens}, the prompt tokens not served from cache plus the completion tokens, which approximates the compute actually purchased for the run. We report per-run medians, and cost as a mean across runs excluding the spend on judging.

None of these quantities is comparable across harness families without qualification, so we treat each as an index of effort rather than of efficiency. An agent turn is not a common unit, because harnesses batch differently: a single step may carry many tool calls in one harness and roughly one in another. Typed tool calls are undercounted for harnesses that reach the tool server through a shell or a code-execution step, since those calls are never typed. Work-tokens depend on what a serving path reports about cache reads, and a path reporting none inflates the figure; where the cache correction must be estimated we give a range rather than a point value. We therefore state the provenance of each figure and compare within a serving path rather than across accounting regimes.
\section{Experimental Results}
\label{sec:experiments}

We evaluate sixteen systems: ten production agent CLIs on their native vendor models, and six open-weight models on a single model-agnostic harness. Using the metrics of Section~\ref{sec:eval_framework}, the experiments target three dissociations the benchmark is built to expose. First, raw accuracy can mask process defects because a correct verdict may be reached through a prohibited shortcut. Second, systems systematically under-abstain, committing to a definitive answer where the record warrants deferral. Third, verdict accuracy and grounding quality dissociate: models reach the right verdict without citing the evidence that justifies it. 

\subsection{Experimental Setup}
\label{sec:exp-setup}

\paragraph{Harnesses and models.} We run four agent harnesses. \texttt{Claude Code}, \texttt{Codex} and \texttt{Gemini CLI} are production agent CLIs, each locked to its own vendor model family and driving the benchmark tools through a native tool-calling loop, so they represent how each model is actually deployed. \texttt{opencode} is a model-agnostic open-source harness used to drive each of the six open-weight models. All four connect to the MIMIC-IV environment through an MCP interface and run with web search and fetch disabled, so every system answers from the in-environment patient record and policy corpus alone. MIMIC data is reached only through a credential-isolated tool server, so no system can bypass the governed tools to read raw records. Table~\ref{tab:exp-matrix} lists the sixteen systems evaluated. Section~\ref{sec:ablations} separately evaluates paired changes to the harness and model configuration.

\begin{table}[htbp]
\centering
\small
\setlength{\tabcolsep}{6pt}
\renewcommand{\arraystretch}{1.2}
\caption{\textbf{Systems evaluated in CliniCARE-Bench.} Native leaderboard (each harness on its own frontier model) and the open-weight model comparison on the neutral \texttt{opencode} harness.}
\label{tab:exp-matrix}
\begin{tabularx}{\linewidth}{@{}lX@{}}
\toprule
\textbf{Harness} & \textbf{Models} \\
\midrule
\multicolumn{2}{@{}l}{\textit{Native leaderboard: each harness on its own frontier model}} \\
\addlinespace[2pt]
\texttt{Claude Code} & Claude Opus 5, Claude Sonnet 5 \\
\texttt{Codex}       & GPT-5.6 Sol, GPT-5.6 Luna, GPT-5.5, GPT-5.4, GPT-5.4-mini \\
\texttt{Gemini CLI}  & Gemini 3.1 Pro, Gemini 3.5 Flash, Gemini 3.6 Flash \\
\addlinespace[6pt]
\midrule
\multicolumn{2}{@{}l}{\textit{Controlled comparison: open-weight models on one neutral harness}} \\
\addlinespace[2pt]
\texttt{opencode} & DeepSeek V4 Pro, DeepSeek V4 Flash, Qwen 3.7 Plus, GLM 5.2,
                    MiniMax M3, Kimi K2.7 Code \\
\bottomrule
\end{tabularx}
\end{table}

\paragraph{Metrics.} Unless noted otherwise, all metrics follow Section~\ref{sec:eval_framework}. We abbreviate the two indeterminate classes, Lack of Data and Medical Ambiguity, as \textit{lack} and \textit{amb} respectively. Accuracy is reported per run (micro); for a system evaluated on all 750 cases, this equals the scenario-macro average, but the two differ for partial runs. We additionally report expected calibration error (ECE), computed from the stated confidence included in each parseable report and defined in Eq.~\ref{eq:ece}.

\subsection{Results}
\label{sec:main-results}

Table~\ref{tab:main750} reports the primary leaderboard on all 750 cases. We analyze the three column groups in turn below, quality, abstention \& calibration, and resource use \& cost.

\begin{table}[htbp]
\centering
\caption{\textbf{Primary leaderboard.} Each system is evaluated once per case. \textit{Acc.} is four-class verdict accuracy. \emph{Def-free} is defect-free accuracy and \emph{Gap} the defect gap over report-present runs. \emph{Proc} is the process rubric score, \emph{Over-com}$/$\emph{Over-abs} the directional abstention errors, and \emph{ECE} expected calibration error. All values are percentages except \emph{Mac-F1} and \emph{ECE}, which are on a 0–1 scale, and Gap, which is in percentage points. Cost columns are per-run medians except USD$/$run (a mean, excluding judge spend). \texttt{opencode} work-token and USD figures are estimated ranges.}
\label{tab:main750}
\footnotesize
\setlength{\tabcolsep}{2.2pt}
\begin{tabular}{lcccccccccccc}
\toprule
 & \multicolumn{5}{c}{\textbf{Quality}} & \multicolumn{3}{c}{\textbf{Abstention \& calibration}} & \multicolumn{4}{c}{\textbf{Resource use \& cost}} \\
\cmidrule(lr){2-6}\cmidrule(lr){7-9}\cmidrule(lr){10-13}
\textbf{System} & \textbf{Acc} & \textbf{Mac-F1} & \textbf{Def-free} & \textbf{Gap} & \textbf{Proc} & \textbf{Over-com} & \textbf{Over-abs} & \textbf{ECE} & \textbf{Turns} & \textbf{Tools} & \textbf{Work-tok} & \textbf{USD/run} \\
\midrule
\multicolumn{13}{@{}l}{\texttt{Claude Code}}\\
\quad Opus 5 & 75.6 & 0.701 & 70.0 & 5.6 & 82.0 & 32.0 & 10.5 & 0.064 & 17 & 28 & 130.3k & 2.05 \\
\quad Sonnet 5 & 73.9 & 0.654 & 65.6 & 8.3  & 70.8 & 30.7 & 10.8 & 0.046 & 13 & 22 & 90.7k & 0.92 \\
\addlinespace
\multicolumn{13}{@{}l}{\texttt{Codex}}\\
\quad GPT-5.6-Sol      & 73.2 & 0.664 & 67.5 & 5.7 & 80.0 & 31.3 & 12.0 & 0.194 & 22 & 16 & 88.2k & 1.12 \\
\quad GPT-5.6-Luna     & 66.8 & 0.588 & 60.8 & 6.0 & 75.2 & 39.3 & 16.5 & 0.229 & 22 & 18 & 98.1k & 0.21 \\
\quad GPT-5.5 & 76.1 & 0.699 & 71.3 & 4.8 & 78.6 & 38.7 & 7.7 & 0.090 & 18 & 48 & 121.9k & 1.51 \\
\quad GPT-5.4          & 71.1 & 0.632 & 65.6 & 5.5 & 76.1 & 36.7 & 11.2 & 0.153 & 21 & 52 & 109.0k & 0.73 \\
\quad GPT-5.4-mini     & 66.4 & 0.559 & 56.1 & 10.3 & 65.2 & 52.7 & 12.2 & 0.215 & 24 & 42 & 84.4k & 0.20 \\
\addlinespace
\multicolumn{13}{@{}l}{\texttt{Gemini CLI}}\\
\quad Gemini-3.6-Flash & 72.9 & 0.652 & 59.1 & 13.9 & 72.6 & 38.0 & 7.3 & 0.251 & 62 & 37 & 389.1k & 1.15 \\
\quad Gemini-3.5-Flash & 71.9 & 0.622 & 60.0 & 11.9 & 71.1 & 42.0 & 7.5 & 0.262 & 62 & 37 & 384.9k & 1.17 \\
\quad Gemini-3.1-Pro   & 72.7 & 0.664 & 57.9 & 14.8 & 56.4 & 21.3 & 11.2 & 0.256 & 42 & 23 & 190.2k & 0.95 \\
\addlinespace
\multicolumn{13}{@{}l}{\texttt{opencode} \textnormal{(open-weight models on the neutral harness)}}\\
\quad DeepSeek-V4-Pro & 68.7 & 0.612 & 58.7 & 10.0 & 69.4 & 40.7 & 9.7 & 0.192 & 14 & 40 & 63--582k & 0.19--1.11 \\
\quad DeepSeek-V4-Flash & 70.7 & 0.586 & 58.3 & 12.4 & 68.6 & 52.0 & 7.8 & 0.219 & 15 & 46 & 79--746k & 0.03--0.12 \\
\quad GLM-5.2 & 73.6 & 0.643 & 64.8 & 8.8  & 76.4 & 42.0 & 7.5 & 0.130 & 14 & 31 & 66--605k & 0.20--1.08 \\
\quad Qwen-3.7-Plus & 65.3 & 0.579 & 53.1 & 12.3 & 65.9 & 46.0 & 14.0 & 0.187 & 11 & 28 & 47--434k & 0.05--0.19 \\
\quad MiniMax-M3 & 65.9 & 0.555 & 54.7 & 11.2 & 72.5 & 55.3 & 10.5 & 0.191 & 21 & 45 & 107--995k & 0.11--0.39 \\
\quad Kimi-K2.7-Code & 66.5 & 0.558 & 54.1 & 12.4 & 69.6 & 52.7 & 10.5 & 0.156 & 20 & 55 & 130--1245k & 0.91--3.27 \\
\bottomrule
\end{tabular}
\end{table}

\paragraph{No system exceeds 76.1\% accuracy on the cohort.} Accuracy ranges from 65.3\% to 76.1\% and macro-F1 ranges from 0.555 to 0.701. Macro-F1 is lower than accuracy for every system because it gives equal weight to the two less frequent \textit{Indeterminate} classes, where performance is weakest. The systems form a continuum rather than distinct tiers: no adjacent systems differ by more than 2.0 percentage points across the 10.8-point accuracy range, and the harness families interleave throughout the ranking. GPT-5.4-mini (66.4\%) and GPT-5.6-Luna (66.8\%) fall below four and three open-weight systems, respectively. The strongest open-weight system, GLM-5.2 at 73.6\%, is 2.5 points behind the leading system, GPT-5.5 at 76.1\%. Because these differences are small, formal comparisons should use paired case-level tests rather than marginal standard errors for individual accuracies.

\begin{figure}[t]
\centering
\includegraphics[width=\linewidth]{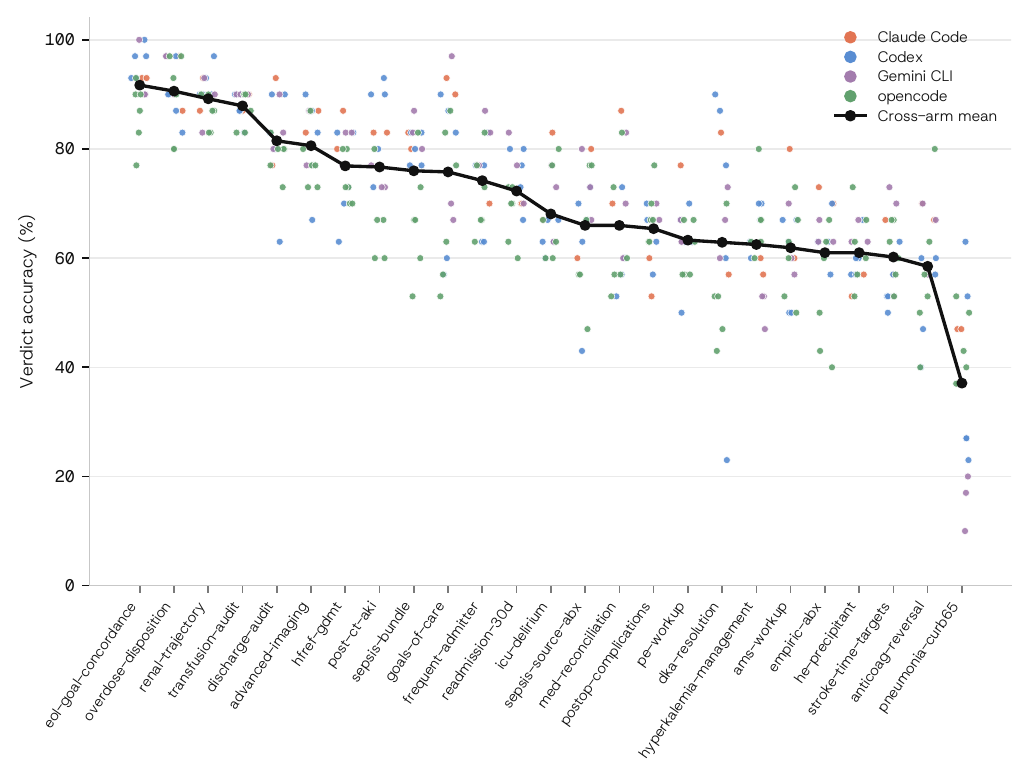}
\caption{\textbf{The leaderboard order does not hold scenario by scenario.} Verdict accuracy of every system on each of the 25 scenarios, one dot per system colored by harness family, with the cross-system mean (dark line). Scenarios are sorted by that mean. Numbers reported in Table~\ref{tab:scenario-table}.}
\label{fig:scenario}
\end{figure}

\paragraph{Scenario by scenario, ranking is unstable and difficulty is concentrated.} Resolved per scenario (\Cref{fig:scenario}), cohort ordering is reversed in 30.2\% of the 3,000 system-pair × scenario comparisons and tied in a further 12.7\%, where 30 cases per scenario cannot resolve the difference. Fourteen of the sixteen systems are the best system on at least one scenario, including Kimi-K2.7-Code, fourth from last on the leaderboard, which leads \texttt{he-precipitant} outright and ties for the lead on \texttt{overdose-disposition}. The ordering of the averages is nevertheless stable: a system's cohort accuracy predicts its mean per-scenario rank almost perfectly (Spearman $\rho=-0.94$, negative because rank 1 is best). The leaderboard is thus a statement about a case mix more than about which system to trust on a given clinical question.

Furthermore, scenario difficulty is concentrated in a few cases: cross-system mean accuracy runs from 91.7\% on \texttt{eol-goal-concordance} to 37.1\% on \texttt{pneumonia-curb65}, and the five hardest scenarios carry 30\% of all errors against the 20\% that uniform difficulty implies. Still, systems range from 10\% to 63\% on the hardest scenario, a 53 point spread, and the widest spread in the grid belongs to \texttt{dka-resolution} (23\% to 90\%), a scenario of middling average difficulty, suggesting that difficulty and discrimination are not the same axis.

\paragraph{Defect-free accuracy reorders the field.} Defect-free accuracy counts a report-present run as correct only when the verdict is right \emph{and} no \textsc{must-not} criterion is violated. Compared to raw accuracy on report-present runs, every system loses between 4.8 and 14.8 points, corresponding to roughly one in sixteen to one in five correct verdicts failing the process audit.

\begin{figure}[t]
\centering
\includegraphics[width=\linewidth]{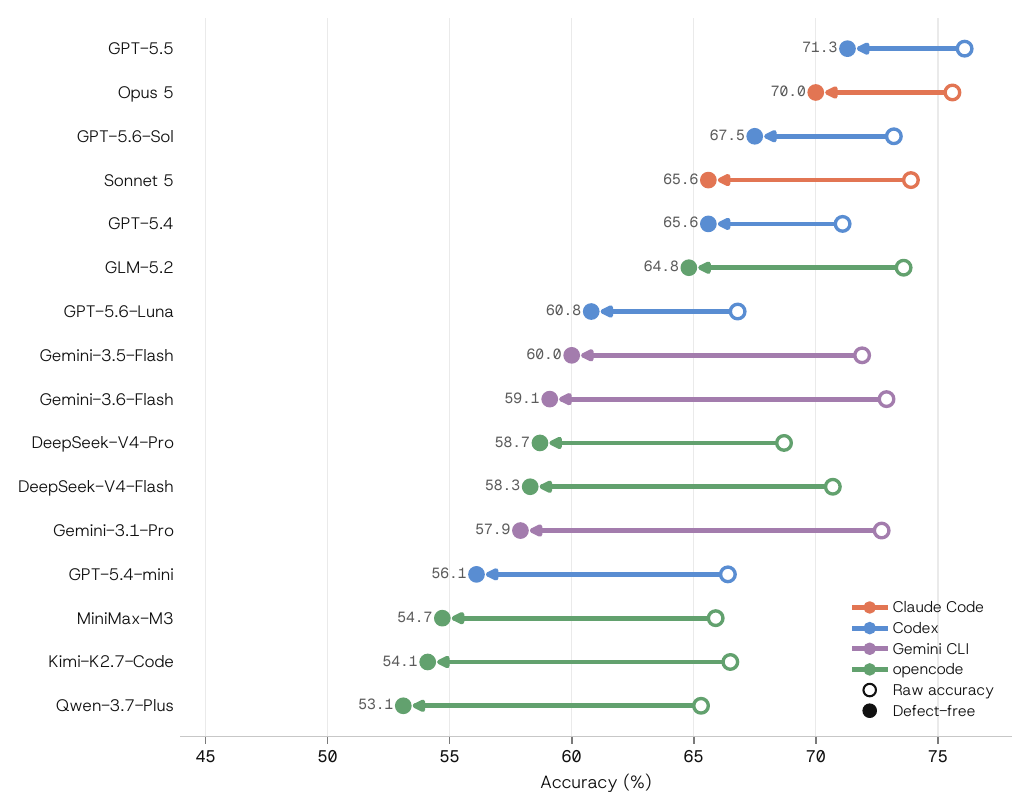}
\caption{\textbf{Defect-free accuracy reorders the leaderboard.} Raw accuracy (open circles) and defect-free accuracy (filled markers) for all sixteen systems, ordered by defect-free accuracy; marker color denotes the harness family and connector length is the defect gap. Gemini-3.1-Pro falls from within noise of the leaders to below every Claude and Codex system bar GPT-5.4-mini once forbidden-route answers are removed, while GPT-5.5 loses least.}
\label{fig:trap}
\end{figure}

The spread is the finding, not the magnitude (\Cref{fig:trap}). Were raw accuracy merely optimistic by a fixed margin, the gap would be roughly constant and the ranking would survive the correction. It does not. Gemini-3.1-Pro reaches 72.7\% raw accuracy, within noise of the leading group, then falls to 57.9\% defect-free, below every Claude system and every Codex system except GPT-5.4-mini, which it clears by 1.8 points. The reordering also promotes: GPT-5.6-Sol trails Sonnet 5 by 0.7 points on raw accuracy but leads it by 1.9 points once forbidden routes are removed. This operationalizes, on real longitudinal records, the outcome--process gap documented for medical LLMs \cite{bedi2025fidelity, natmed2026robustness}: a correct label is not evidence of correct reasoning, and only a trace-level audit of the investigation distinguishes the two.

\paragraph{The process score measures something the verdict does not.} Process adherence separates systems that accuracy places together. Opus 5 scores 82.0\% on process against 75.6\% accuracy, whereas Gemini-3.1-Pro reaches 72.7\% accuracy on a process score of 56.4\%, the same verdict quality reached through a materially less defensible investigation. This is the defect gap seen from the other side. For a clinical deliverable this matters more than the headline: a reader can audit a documented investigation but not a bare verdict.

\begin{figure}[t]
\centering
\includegraphics[width=\linewidth]{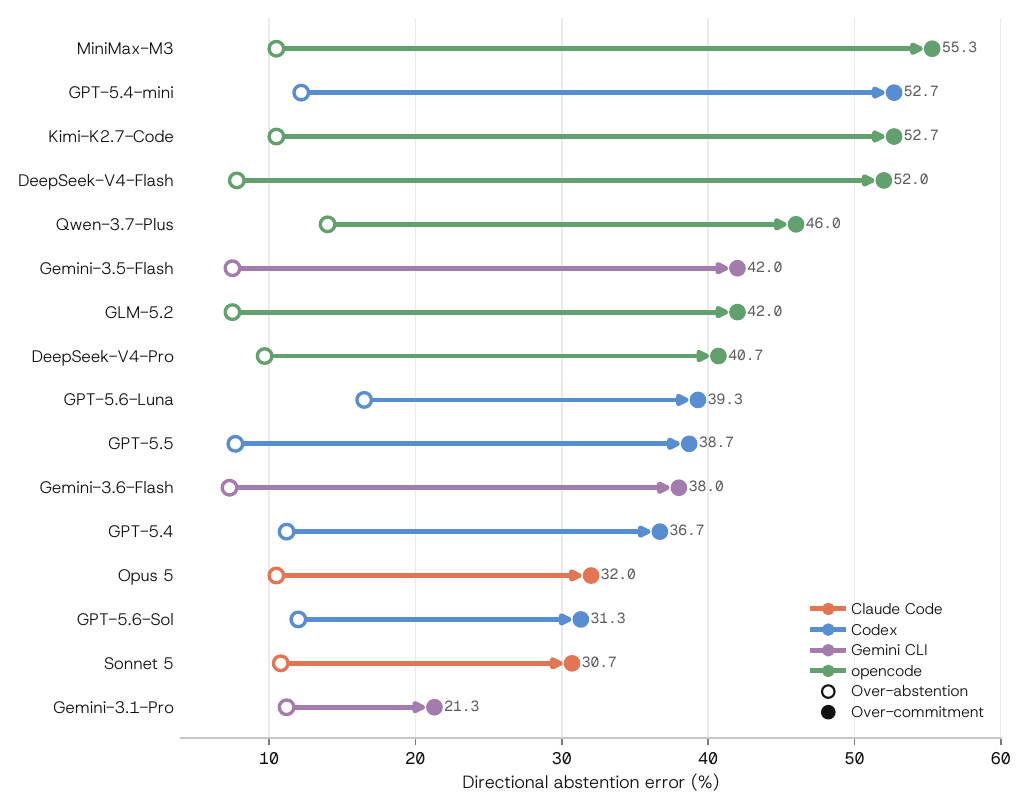}
\caption{\textbf{Every system under-abstains.} Directional abstention errors for all sixteen systems: over-abstention (open circles), computed over reference-definitive runs, and over-commitment (filled markers), computed over reference-abstention runs; marker color denotes the harness family. Over-commitment exceeds over-abstention in every system. GPT-5.4-mini, DeepSeek-V4-Flash, MiniMax-M3 and Kimi-K2.7-Code over-commit on at least half of all deferral cases, whereas Gemini-3.1-Pro shows the smallest directional gap.}
\label{fig:abstention}
\end{figure}

\begin{figure}[t]
\centering
\includegraphics[width=\linewidth]{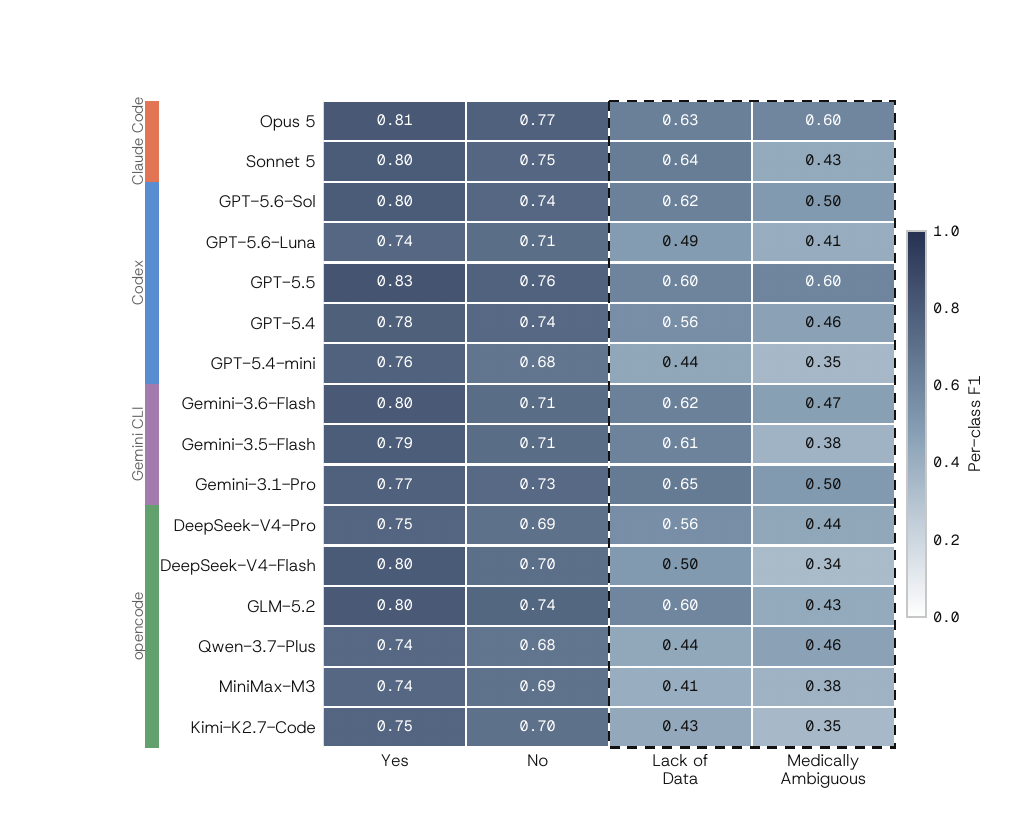}
\caption{\textbf{Performance collapses on the two safety-critical \textit{Indeterminate} classes.} Per-class F1 for all sixteen systems on the complete cohort (rows in leaderboard order, grouped by harness via the color bar at left). \textit{Yes} and \textit{No} stay strong while both deferral classes (dashed box) fall away on the weaker systems. Precision on the deferral classes is low throughout (\textit{lack} 46--67\%, \textit{amb} 26--65\%), as seen in Table~\ref{tab:main750-perclass}.}
\label{fig:perclass}
\end{figure}

\paragraph{Systems under-abstain: over-commitment exceeds over-abstention in every system.} Measured directionally (\Cref{fig:abstention}), over-commitment runs 21.3--55.3\% against over-abstention's 7.3--16.5\%, and \emph{all sixteen} systems over-commit at the higher rate. That the asymmetry holds system by system rather than only in aggregate makes it a property of the systems rather than of the label distribution, and it reproduces on the development subset (Section~\ref{sec:ablations}). The magnitude, not the direction, is what varies: four systems over-commit on at least half of all deferral cases (MiniMax-M3 55.3\%, DeepSeek-V4-Flash 52.0\%, GPT-5.4-mini and Kimi-K2.7-Code 52.7\%) while Gemini-3.1-Pro does so on 21.3\%. Over-abstention, by contrast, is tightly banded at 7.3--16.5\%, and its highest value belongs to a vendor system (GPT-5.6-Luna), so the deferral side of the error budget varies far less across systems than the commitment side. Breaking accuracy down by class shows that even when models do abstain, they often do so on records that in fact settle the question (\Cref{fig:perclass}): \textit{Yes} and \textit{No} F1 stay strong across the roster while both \textit{Indeterminate} classes fall away, and precision on the deferral classes is low throughout (\textit{lack} 46--67\%, \textit{amb} 26--65\%), as seen in Table~\ref{tab:main750-perclass}.

\paragraph{Calibration separates the systems more cleanly than accuracy does.} Every parseable report includes a stated confidence in its class prediction, so we can ask whether that number carries information. Expected calibration error is the bin-weighted deviation between stated confidence and realized accuracy,
\begin{equation}
\mathrm{ECE} \;=\; \sum_{b=1}^{B} \frac{|B_b|}{n}\,\bigl|\,\mathrm{conf}(B_b) - \mathrm{acc}(B_b)\,\bigr| ,
\label{eq:ece}
\end{equation}
over $B=10$ equal-width bins of stated confidence, where $n$ is the number of report-present runs with a valid confidence and $\mathrm{conf}(B_b)$ and $\mathrm{acc}(B_b)$ are the mean stated confidence and realized accuracy in bin $b$. ECE spans 0.046--0.262 across the sixteen systems, a nearly sixfold range on an axis accuracy cannot see, and it does \emph{not} order them the way accuracy does: the three best-calibrated are Sonnet 5 (0.046), Opus 5 (0.064) and GPT-5.5 (0.090), while the three worst are all Gemini CLI systems (0.251--0.262) which sit mid-field on accuracy. In particular, the Gemini systems all emit only a few distinct confidence values, and their ECE reflects a lack of graded uncertainty rather than a poorly-shaped calibration curve. For a report addressed to a clinician who cannot cheaply re-derive the answer, a confidence that does not track correctness is a more consequential defect than a few points of accuracy.

\begin{figure}[t]
\centering
\includegraphics[width=\linewidth]{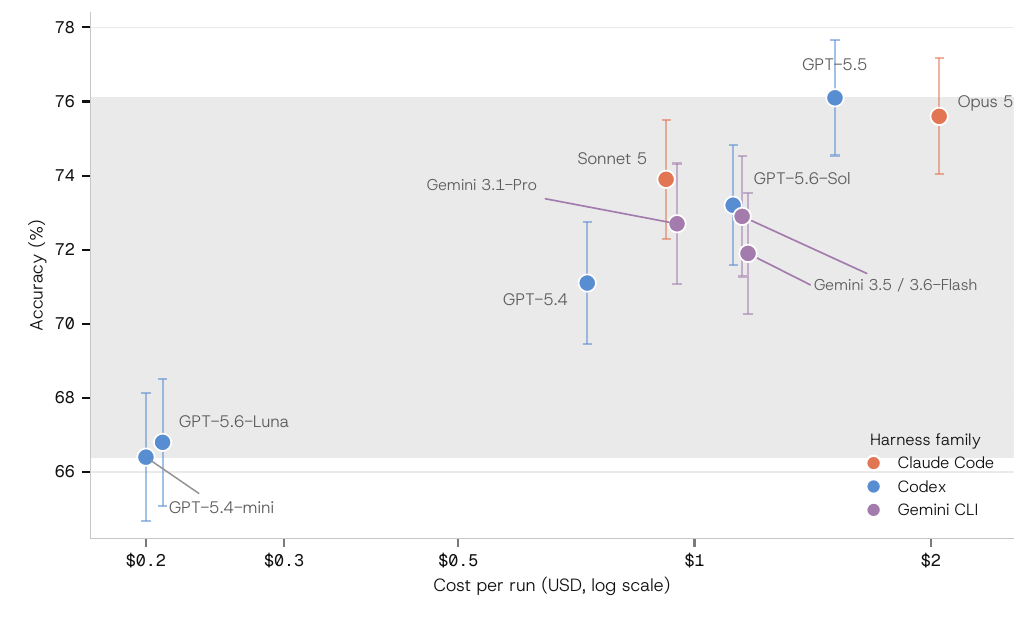}
\caption{\textbf{Accuracy rises only weakly with cost.} Per-run cost (USD, log scale, harness-reported) against verdict accuracy for the ten vendor systems, colored by harness family, with $\pm1$ binomial standard error bars at $n=750$. The six \texttt{opencode} systems are omitted because their per-run cost is a range rather than a point.}
\label{fig:cost}
\end{figure}

\paragraph{Resource use and cost diverge from accuracy, and from each other.} Across the vendor systems, a tenfold spread in per-run cost corresponds to only an approximately 10 point accuracy range (\Cref{fig:cost}). GPT-5.4-mini and GPT-5.6-Luna cost nearly the same (\$0.20 and \$0.21 per run) and differ by only 0.4 points in accuracy, yet GPT-5.6-Luna achieves 10.0 points higher process adherence and 4.7 points higher defect-free accuracy. Because the systems are similarly priced, these differences cannot be attributed to additional spend. Instead, they show that systems with comparable cost and verdict accuracy can differ substantially in the defensibility of their investigations. Their resource profiles also differ: GPT-5.4-mini issues a median of 42 typed tool calls compared with GPT-5.6-Luna's 18, roughly $2.3\times$ as many, while consuming fewer work-tokens (84.4k versus 98.1k). The system making more typed tool calls nevertheless has the higher defect gap and lower process score, showing that greater retrieval activity alone does not ensure sound adjudication of the retrieved evidence.

\subsection{Qualitative Analysis}
\label{sec:qualitative}
We present an analysis of one case in detail to illustrate some of the failure signatures measured above. A cirrhotic patient is admitted with hepatic encephalopathy and moderate ascites, and no diagnostic paracentesis appears anywhere in the admission, so spontaneous bacterial peritonitis can be neither confirmed nor excluded and the reference verdict is \textit{Indeterminate: Lack of Data} (\texttt{he-precipitant-278ac1}). Thirteen of the sixteen systems commit to a definitive label, twelve of them \textit{Yes}, at stated confidences from 78\% to 100\%. Seven reach that label through the same prohibited shortcut, asserting peritonitis excluded with no ascitic-fluid result, the \textsc{must-not} criterion N4 (Table~\ref{tab:qual-he-precipitant}). The shortcut is not confined to one family: it is sprung by one Claude system, all three Gemini systems, and three of the six open-weight systems, while no Codex system encounters it. Not one of the sixteen earns D2, the criterion requiring confirmation that the major candidates, paracentesis especially, were actually evaluated.

Two systems make the separation of process from verdict concrete, in opposite directions. GPT-5.6-Sol is the only system in the roster to earn D5, which rewards escalating a key candidate that was never worked up despite indication; it springs no defect and carries the strongest \textsc{must-do} profile of any committing system, and it still emits a definitive \textit{No}. Recognizing an evidentiary gap is therefore not sufficient: the failure occurs at label assignment, after the reasoning that should have prevented it. Running the other way, the three systems that return the correct \textit{Indeterminate} verdict --- DeepSeek-V4-Pro, GLM-5.2 and Qwen-3.7-Plus --- earn \emph{none} of the nine process criteria between them, and one of the three states 95\% confidence in it. Their label is right and their investigation is undocumented. A single case thus exhibits both failure modes the two axes are designed to distinguish: sound reasoning that does not produce the correct label, and a correct label that reflects no defensible process. Scoring either axis alone would record one of these systems as exemplary and the other as broken, when the trace shows neither is.

\begin{table}[htbp]
\centering
\small
\caption{\textbf{Qualitative failure profile} (\texttt{he-precipitant-278ac1}, all sixteen systems).
Thirteen of sixteen commit to a definitive label; seven reach it through the same \textsc{must-not} criterion N4,
``Asserted SBP excluded without an ascitic-fluid result''. \textsc{must-do} criteria are D1--D5 (D5:
``Escalated when a key candidate was not worked up despite indication''). GPT-5.6-Sol is the only
system to earn D5 and still commits; the three systems that return the correct verdict earn none of
the nine criteria.}
\label{tab:qual-he-precipitant}
\begin{tabular}{llccl}
\toprule
\textbf{System} & \textbf{Verdict} & \textbf{Stated conf.} & \textbf{\textsc{must-do} met} & \textbf{\textsc{must-not} sprung} \\
\midrule
\emph{Gold} & \textit{Ind.: Lack of Data} & -- & -- & -- \\
\midrule
Claude Opus 5 (Claude Code)      & Yes                          & \phantom{0}85\%  & D3,D4        & -- \\
Claude Sonnet 5 (Claude Code)    & Yes                          & \phantom{0}90\%  & D1,D3,D4     & N4 \\
GPT-5.6-Sol (Codex)              & No                           & \phantom{0}88\%  & D3,D4,\textbf{D5} & -- \\
GPT-5.6-Luna (Codex)             & Yes                          & \phantom{0}88\%  & D3,D4        & -- \\
GPT-5.5 (Codex)                  & Yes                          & \phantom{0}82\%  & D1,D3,D4     & -- \\
GPT-5.4 (Codex)                  & Yes                          & \phantom{0}78\%  & D3,D4        & -- \\
GPT-5.4-mini (Codex)             & Yes                          & \phantom{0}92\%  & D3,D4        & -- \\
Gemini-3.6-Flash (Gemini CLI)    & Yes                          & \phantom{0}95\%  & D3,D4        & N4 \\
Gemini-3.5-Flash (Gemini CLI)    & Yes                          & 100\%            & D3,D4        & N4 \\
Gemini-3.1-Pro (Gemini CLI)      & Yes                          & \phantom{0}95\%  & D3           & N4 \\
DeepSeek-V4-Pro (opencode)       & \textit{Ind.: Lack of Data}  & \phantom{0}0\%               & --           & -- \\
DeepSeek-V4-Flash (opencode)     & Yes                          & \phantom{0}95\%  & D3           & N4 \\
GLM-5.2 (opencode)               & \textit{Ind.: Lack of Data}  & \phantom{0}95\%  & --           & -- \\
Kimi-K2.7-Code (opencode)        & Yes                          & \phantom{0}85\%  & D3           & N4 \\
MiniMax-M3 (opencode)            & Yes                          & \phantom{0}85\%  & D3,D4        & N4 \\
Qwen-3.7-Plus (opencode)         & \textit{Ind.: Lack of Data}  & \phantom{0}50\%  & --           & -- \\
\bottomrule
\end{tabular}
\end{table}

\subsection{Patient-Evidence Grounding}
\label{sec:grounding-results}

Grounding is where verdict accuracy and defensible reasoning most clearly diverge: a system can reach the correct verdict without citing the evidence that justifies it. Table~\ref{tab:grounding-v3-full750} reports the four patient-evidence grounding metrics of Section~\ref{sec:patient-grounding} --- $P_{\mathrm{record}}$ and $P_{\mathrm{claim}}$ for citation precision, $R_{\mathrm{retrieval}}$ and $R_{\mathrm{finding}}$ for evidence coverage --- for all sixteen model--harness combinations over the same 750 cases. All four are computed against the agent's own investigation rather than against the reference verdict, so they capture whether a determination is auditable, not whether it is correct. All four are stable under repetition: $P_{\mathrm{record}}$ and $R_{\mathrm{retrieval}}$ are rule-based and return the same value on every computation, and both $P_{\mathrm{claim}}$ and $R_{\mathrm{finding}}$ reproduce reliably when the same reports are graded again (Section~\ref{sec:judge-validation}).

\begin{table}[htbp]
\centering
\caption{\textbf{Patient-evidence grounding on all 750 cases.} Citation precision and evidence coverage follow Section~\ref{sec:patient-grounding}. Values are averaged over grounding-scoreable runs, defined as report-present runs in which the system retrieved at least one patient record.}
\label{tab:grounding-v3-full750}
\footnotesize
\setlength{\tabcolsep}{8pt}
\begin{tabular}{lcccc}
\toprule
 & \multicolumn{2}{c}{\textbf{Citation precision}} & \multicolumn{2}{c}{\textbf{Evidence coverage}} \\
\cmidrule(lr){2-3}\cmidrule(lr){4-5}
\textbf{System} & $P_{\mathrm{record}}$ & $P_{\mathrm{claim}}$ & $R_{\mathrm{retrieval}}$ & $R_{\mathrm{finding}}$ \\
\midrule
\multicolumn{5}{@{}l}{\texttt{Claude Code}}\\
\quad Opus 5           & 0.985 & 0.884 & 0.852 & 0.843 \\
\quad Sonnet 5         & 0.988 & 0.869 & 0.818 & 0.744 \\
\addlinespace
\multicolumn{5}{@{}l}{\texttt{Codex}}\\
\quad GPT-5.6-Sol      & 1.000 & 0.971 & 0.848 & 0.768 \\
\quad GPT-5.6-Luna     & 1.000 & 0.965 & 0.871 & 0.737 \\
\quad GPT-5.5          & 0.998 & 0.964 & 0.878 & 0.763 \\
\quad GPT-5.4          & 0.999 & 0.961 & 0.864 & 0.738 \\
\quad GPT-5.4-mini     & 0.997 & 0.964 & 0.813 & 0.673 \\
\addlinespace
\multicolumn{5}{@{}l}{\texttt{Gemini CLI}}\\
\quad Gemini-3.6-Flash & 0.995 & 0.899 & 0.859 & 0.789 \\
\quad Gemini-3.5-Flash & 0.996 & 0.893 & 0.863 & 0.790 \\
\quad Gemini-3.1-Pro   & 0.999 & 0.903 & 0.689 & 0.604 \\
\addlinespace
\multicolumn{5}{@{}l}{\texttt{opencode} \textnormal{(open-weight models on the neutral harness)}}\\
\quad DeepSeek-V4-Pro & 0.997 & 0.908 & 0.873 & 0.764 \\
\quad DeepSeek-V4-Flash & 0.989 & 0.917 & 0.877 & 0.757 \\
\quad GLM-5.2 & 0.993 & 0.925 & 0.842 & 0.785 \\
\quad Qwen-3.7-Plus & 0.994 & 0.930 & 0.821 & 0.732 \\
\quad MiniMax-M3 & 0.992 & 0.912 & 0.855 & 0.778 \\
\quad Kimi-K2.7-Code & 0.997 & 0.940 & 0.900 & 0.767 \\
\bottomrule
\end{tabular}
\end{table}

\paragraph{Record-identifier precision saturates and does not separate the systems.} $P_{\mathrm{record}}$ spans 0.985--1.000 across the systems. No system fabricates record identifiers at a rate this check resolves, so the column establishes a lower bound on citation validity rather than discriminating between systems.

\paragraph{Claim-level precision tracks the harness, not model capability.} $P_{\mathrm{claim}}$ separates along harness families --- Codex 0.965, \texttt{opencode} 0.922, Gemini CLI 0.898, Claude Code 0.877 --- while within-family variation is much smaller. Each vendor harness holds its models tightly together despite spanning that vendor's range from small to frontier, so capability differences large enough to reorder verdict accuracy leave claim-level precision essentially unchanged. The six unrelated open-weight models on \texttt{opencode} are the one exception: their spread is 0.03 compared to within-family spreads of 0.01--0.015, but this remains a fraction of the gap separating the strongest and weakest family. Across the sixteen systems the $P_{\mathrm{claim}}$ ranking shows no positive association with the verdict-accuracy ranking of Table~\ref{tab:main750} (Spearman $\rho=-0.28$). This is the mirror image of the harness ablation (Table~\ref{tab:ablation-harness}), where varying the harness at a fixed model left verdict accuracy inside noise: whether a determination is traceable to retrieved evidence is set largely by the scaffold that structures retrieval and report generation, whereas whether it is correct is set by the model.

\paragraph{Citation precision and evidence coverage are weakly anti-correlated.} Across the 16 systems, $P_{\mathrm{claim}}$ correlates negatively with $R_{\mathrm{finding}}$ ($r=-0.22$) and weakly with $R_{\mathrm{retrieval}}$ ($r=+0.23$), while the two coverage metrics correlate strongly ($r=+0.77$). This precision-coverage trade-off characterizes the field broadly but is most pronounced at the family level. For example, Codex and Claude Code occupy opposite extremes, where the most precise family is the least complete, and vice versa. However, this constraint does not dictate individual system performance; \texttt{opencode}, for instance, ranks second on both axes. Claim classifications reveal the mechanism behind this trade-off: across all families, ungrounded claims are predominantly \emph{unverifiable} rather than \emph{contradicted} (ranging from 3:1 to 5:1). Systems achieving higher coverage typically over-extrapolate beyond their retrieved evidence rather than misread it.

\subsection{Policy Grounding}
\label{sec:policy-results}

Policy grounding is scored separately from patient-evidence grounding (Section~\ref{sec:policy-grounding-metrics}). We report its two axes: whether policy citations resolve and support the claims they carry, and whether the report identifies the governing documents, together with the effect of corpus access itself (Tables~\ref{tab:main750-policy} and~\ref{tab:ablation-policy}). Comparing the policy-on and policy-off conditions isolates whether corpus access changes system behavior.

\paragraph{Systems cite the governing policy correctly but incompletely.} Policy grounding on the cohort (Table~\ref{tab:main750-policy}) separates two failure modes a single grounding score would merge. Take GPT-5.4-mini and GPT-5.6-Luna, two systems complete at 750 and within 0.4 accuracy points of each other. Citations almost always resolve (0.995 and 0.999), and precision of the cited document set against the clinical board's governing list is high (0.800 and 0.825). Recall of that set is roughly half (0.479 and 0.524): systems cite documents that genuinely govern, then omit about half of the others that also govern. Since a determination resting on part of the applicable standard can be right for the wrong reason, this is the policy-axis analogue of the defect gap, and a combined score would hide it because high precision offsets low recall. That failure shape is the one finding here that reproduces across all sixteen systems, because these are rates rather than counts and each is computed within a system: resolution runs 0.954--1.000 and document precision 0.764--0.899, while document recall never exceeds 0.703 and falls as low as 0.407. Every system in the roster predominantly cites real, applicable policy documents while omitting a substantial share of the other governing documents.

\paragraph{Unresolved policy citations dissociate sharply between systems of equal accuracy.} The same two systems emit 20 and 3 unresolved policy citations, respectively---references that resolve to no corpus document---a nearly sevenfold difference between systems scoring within half a point of each other on accuracy. Citation support, the LLM-judged check that a cited passage entails the claim attached to it, moves the same way (0.629 against 0.814). The dissociation is far starker elsewhere in the roster: Gemini-3.5-Flash emits 303 unresolved citations and MiniMax-M3 159, against zero for GPT-5.5, GPT-5.4 and Gemini-3.1-Pro. With the defect gap and the confidence margins, this is the third axis on which these two systems are clearly distinguishable while their headline accuracy is not, and it is the one a clinician auditing the report would notice first.

\begin{table}[htbp]
\centering
\caption{\textbf{Policy grounding on all 750 cases.} Resolution checks whether a citation names a corpus document and valid line span; Support is LLM-judged; Doc-P and Doc-R compare cited documents with the expert-specified governing set. Resolution and Support are averaged over runs with at least one policy citation, Doc-P over runs citing at least one corpus document, and Doc-R over all runs, with no-citation runs scoring zero. Unresolved citation counts are reported in the text.}
\label{tab:main750-policy}
\small
\setlength{\tabcolsep}{6pt}
\begin{tabular}{lcccc}
\toprule
\textbf{System} & \textbf{Resolve} & \textbf{Support} & \textbf{Doc-P} & \textbf{Doc-R} \\
\midrule
\multicolumn{5}{@{}l}{\texttt{Claude Code}}\\
\quad Opus 5 & 1.000 & 0.564 & 0.764 & 0.703 \\
\quad Sonnet 5 & 0.993 & 0.579 & 0.823 & 0.443 \\
\addlinespace
\multicolumn{5}{@{}l}{\texttt{Codex}}\\
\quad GPT-5.6-Sol      & 1.000 & 0.735 & 0.814 & 0.581 \\
\quad GPT-5.6-Luna     & 0.999 & 0.814 & 0.825 & 0.524 \\
\quad GPT-5.5 & 1.000 & 0.747 & 0.786 & 0.586 \\
\quad GPT-5.4          & 1.000 & 0.624 & 0.782 & 0.549 \\
\quad GPT-5.4-mini     & 0.995 & 0.629 & 0.800 & 0.479 \\
\addlinespace
\multicolumn{5}{@{}l}{\texttt{Gemini CLI}}\\
\quad Gemini-3.6-Flash & 0.985 & 0.467 & 0.771 & 0.623 \\
\quad Gemini-3.5-Flash & 0.984 & 0.467 & 0.769 & 0.614 \\
\quad Gemini-3.1-Pro   & 1.000 & 0.674 & 0.785 & 0.407 \\
\addlinespace
\multicolumn{5}{@{}l}{\texttt{opencode} \textnormal{(open-weight models on the neutral harness)}}\\
\quad DeepSeek-V4-Pro & 0.997 & 0.621 & 0.870 & 0.525 \\
\quad DeepSeek-V4-Flash & 0.954 & 0.566 & 0.849 & 0.452 \\
\quad GLM-5.2 & 0.995 & 0.584 & 0.820 & 0.576 \\
\quad Qwen-3.7-Plus & 0.998 & 0.559 & 0.899 & 0.443 \\
\quad MiniMax-M3 & 0.976 & 0.471 & 0.817 & 0.530 \\
\quad Kimi-K2.7-Code & 0.998 & 0.613 & 0.818 & 0.531 \\
\bottomrule
\end{tabular}
\end{table}

\subsection{Ablations}
\label{sec:ablations}

The ablations use the frozen 148-case development subset described in Appendix~\ref{app:dev-subset}, which oversamples the two Indeterminate classes. We therefore use it only for paired contrasts and repeated-run analyses; its absolute scores are not comparable with Table~\ref{tab:main750}. Paired ablations reuse Opus 4.8, whereas Appendix~\ref{app:dev-leaderboard} slices the full-cohort runs and therefore reports Opus 5 across per-pair tests and per-class breakdowns. All sixteen systems are complete at $n=148$: trials lost to the execution environment were re-run rather than dropped, so every contrast below is paired over the same tasks. The subset is deliberately \emph{not} used to separate systems from one another --- of the $\binom{16}{2}=120$ pairs, only 11 reach $p<0.05$ under a two-sided exact McNemar test and none survives Holm correction. It does, however, confirm shared behavior: the abstention asymmetry of Section~\ref{sec:main-results} reproduces here on a deliberately different label distribution, with over-commitment exceeding over-abstention in \emph{all sixteen} systems (26.4--60.4\% against 5.3--16.8\%), which is what makes it a property of the systems rather than of the cohort's label mix.


\paragraph{Withholding the policy corpus shows no reliable effect on either accuracy or ambiguity recognition.} With the corpus withheld and nothing else changed, accuracy moves \emph{inconsistently}: two of five systems fall, two rise, and one does not move, and a two-sided exact McNemar test on the paired cases separates \emph{no} system's shift from zero ($p=0.21$--$1.00$). Corpus access is therefore not worth a fixed number of points, and these deltas do not order the systems. Recall on \textit{amb} is no more consistent: two systems fall, two rise and one is unchanged. At a support of 14 those movements are one to three cases apiece ($\pm7.1$ points per case), so the column carries no direction either. We had expected withholding the corpus to cost \textit{amb} recall specifically --- the corpus is what tells an agent that a governing standard exists without settling the case --- and it does not, which is a negative result rather than a measurement we can report as directional. The policy-enabled condition remains canonical. This is an ablation of the environment, not an alternative configuration (Table~\ref{tab:ablation-policy}).

\begin{table}[htbp]
\centering
\caption{\textbf{Policy-corpus ablation.} Conditions are paired by case with web search disabled. $\Delta$ is off minus on, so negative values indicate that withholding the corpus reduced performance. Exact two-sided McNemar tests find no accuracy effect ($p=0.21$-$1.00$). \textit{amb} recall has support 14, where one case changes recall by 7.1 points; its observed differences are also not significant ($p=0.25$-$1.00$) and support neither a direction nor a ranking.}
\label{tab:ablation-policy}
\small
\setlength{\tabcolsep}{5pt}
\begin{tabular}{lccccccc}
\toprule
 & \multicolumn{4}{c}{\textbf{Accuracy}} & \multicolumn{3}{c}{\textbf{Recall \textit{amb} \%}} \\
\cmidrule(lr){2-5}\cmidrule(lr){6-8}
\textbf{Model (harness)} & on & off & $\Delta$ & $p$ & on & off & $\Delta$ \\
\midrule
GPT-5.6-Sol (Codex)         & 70.3 & 70.9 & $+0.7$ & 1.00 & 50.0 & 50.0 & $\pm0.0$ \\
GPT-5.5 (Codex)             & 73.0 & 70.9 & $-2.0$ & 0.66 & 64.3 & 42.9 & $-21.4$ \\
Gemini-3.1-Pro (Gemini CLI) & 77.0 & 72.3 & $-4.7$ & 0.21 & 35.7 & 42.9 & $+7.1$ \\
GPT-5.6-Luna (Codex)        & 59.5 & 62.2 & $+2.7$ & 0.56 & 71.4 & 50.0 & $-21.4$ \\
Opus 4.8 (Claude Code)      & 70.3 & 70.3 & $\pm0.0$ & 1.00 & 50.0 & 57.1 & $+7.1$ \\
\bottomrule
\end{tabular}
\end{table}

\paragraph{Single-attempt accuracy overstates consistent correctness, and the inconsistency is systematic.} Across three separately executed attempts, $\mathrm{pass}\mbox{\textasciicircum}3$ falls 8.3--12.6 points below $\mathrm{avg}@3$, and the residual agreement is on the \emph{same wrong} verdict, 13.5--20.9 points, and for GPT-5.6-Luna 30\% of the cases on which it agreed with itself. Run-to-run variation is small next to that shortfall, with the standard deviation of accuracy across the three attempts at 0.7--4.7 points, so the gap is not attempt noise. High self-consistency is therefore not evidence of reliability: these systems are mostly \emph{stably} wrong rather than randomly wrong, so repeating a query will not surface the error. Reliability is also worst exactly where the benchmark is hardest, as \emph{every} system's lowest $\mathrm{pass}\mbox{\textasciicircum}3$ falls on an \textit{Indeterminate} class (Table~\ref{tab:passk-perclass}).

\begin{table}[htbp]
\centering
\caption{\textbf{Repeated-run reliability, development subset.} 
Each system has three independent attempts on all 148 cases. We report the reliability metrics defined in Section~\ref{sec:reliability}. \textit{Spread} is $\mathrm{pass@3}-\mathrm{pass\mbox{\textasciicircum}3}$, and \textit{Overst.} is $\mathrm{avg@3}-\mathrm{pass\mbox{\textasciicircum}3}$. \textit{Agree-wr} is agreement on an incorrect verdict, so \textit{Agree} equals $\mathrm{pass\mbox{\textasciicircum}3}+\mathrm{Agree\mbox{-}wr}$ up to rounding. Sampling error supports the aggregate pattern, not a ranking among systems.}

\label{tab:passk}
\small
\setlength{\tabcolsep}{2.5pt}
\begin{tabular}{lcccccccrr}
\toprule
\textbf{System} & $n_{\mathrm{com}}$ & \textbf{avg@3} & \textbf{SD} & \textbf{pass\textasciicircum{}3} & \textbf{pass@3} & \textbf{Spread} & \textbf{Overst.} & \textbf{Agree} & \textbf{Agree-wr} \\
\midrule
Opus 4.8 (Claude Code)         & 148 & \textbf{72.7} & 3.7 & \textbf{64.2} & \textbf{82.4} & 18.2 & \phantom{0}8.6 & \textbf{79.7} & 15.5 \\
\addlinespace
GPT-5.5 (Codex)                & 148 & 71.8 & 1.4 & 63.5 & 80.4 & \textbf{16.9} & \textbf{\phantom{0}8.3} & 77.0 & 13.5 \\
GPT-5.6-Sol (Codex)            & 148 & 70.3 & 0.7 & 59.5 & 79.7 & 20.3 & 10.8 & 78.4 & 18.9 \\
GPT-5.6-Luna (Codex)           & 148 & 60.6 & 1.0 & 48.0 & 72.3 & 24.3 & 12.6 & 68.9 & 20.9 \\
\addlinespace
Gemini-3.1-Pro (Gemini CLI)    & 148 & 72.3 & 4.7 & 60.8 & \textbf{82.4} & 21.6 & 11.5 & 76.4 & 15.5 \\
\bottomrule
\end{tabular}
\end{table}

\begin{table}[htbp]
\centering
\footnotesize
\setlength{\tabcolsep}{3pt}
\caption{\textbf{Per-class repeated-run reliability, development subset.} Values use three attempts; class supports are \textsc{yes} 50, \textsc{no} 45, \textsc{lack} 39, \textsc{amb} 14. pass\textasciicircum{}3 and pass@3 follow Section~\ref{sec:reliability}. At an \textsc{amb} support of 14, one case changes a cell by 7.1 points, so the amb values indicate direction only and should not be used to rank systems.}

\label{tab:passk-perclass}
\begin{tabular}{lcccccccccccccccc}
\toprule
& \multicolumn{4}{c}{\textsc{yes}} & \multicolumn{4}{c}{\textsc{no}}
& \multicolumn{4}{c}{\textsc{lack}} & \multicolumn{4}{c}{\textsc{amb}} \\
\cmidrule(lr){2-5} \cmidrule(lr){6-9} \cmidrule(lr){10-13} \cmidrule(lr){14-17}
\textbf{Model (harness)} & $n$ & \^{}3 & @3 & SD & $n$ & \^{}3 & @3 & SD & $n$ & \^{}3 & @3 & SD & $n$ & \^{}3 & @3 & SD \\
\midrule
Opus 4.8 (Claude Code)       & 50 & \textbf{78.0} & \textbf{96.0} & 7.2 & 45 & 60.0 & \textbf{86.7} & 3.4 & 39 & 56.4 & 64.1 & 1.5 & 14 & \textbf{50.0} & 71.4 & 7.1 \\
\addlinespace
GPT-5.5 (Codex)              & 50 & 70.0 & 88.0 & 2.0 & 45 & \textbf{73.3} & 84.4 & 1.3 & 39 & 51.3 & 71.8 & 3.0 & 14 & 42.9 & 64.3 & 10.9 \\
GPT-5.6-Sol (Codex)          & 50 & 68.0 & 88.0 & 1.2 & 45 & 64.4 & 82.2 & 4.4 & 39 & 48.7 & 71.8 & 2.6 & 14 & 42.9 & 64.3 & 4.1 \\
GPT-5.6-Luna (Codex)         & 50 & 52.0 & 72.0 & 5.3 & 45 & 48.9 & 73.3 & 1.3 & 39 & 43.6 & 69.2 & 2.6 & 14 & 42.9 & \textbf{78.6} & 12.4 \\
\addlinespace
Gemini-3.1-Pro (Gemini CLI)  & 50 & 64.0 & 86.0 & 5.3 & 45 & 64.4 & \textbf{86.7} & 8.0 & 39 & \textbf{69.2} & \textbf{84.6} & 3.0 & 14 & 14.3 & 50.0 & 4.1 \\
\bottomrule
\end{tabular}
\end{table}


\paragraph{At a fixed model, the harness is a cost choice, not a capability choice.} Holding the model fixed and varying the harness across three model-agnostic harnesses moves accuracy little: the spread is 4.1 points, smaller than the across-model spread within a single harness, and no pairwise difference approaches significance on a paired test (Table~\ref{tab:ablation-harness}). What separates the harnesses is cost, as median completion tokens differ 4.3-fold for accuracy differences inside noise, so for this model the harness is primarily a resource-use choice rather than a clearly separable capability difference.

\begin{table}[htbp]
\centering
\caption{\textbf{Harness comparison at a fixed model.} DeepSeek-V4-Pro is evaluated with three model-agnostic harnesses on the same 148 tasks. Accuracy differs by at most 4.1 points (exact McNemar $p\geq0.46$), while median completion tokens differ 4.3-fold. The three builds carry identical task-content digests.}
\label{tab:ablation-harness}
\small
\setlength{\tabcolsep}{5pt}
\begin{tabular}{lcccccc}
\toprule
\textbf{Harness} & \textbf{Acc \%} & $\Delta$ & $p$ & \textbf{Proc\%} & \textbf{Turns} & \textbf{Comp.\ tok} \\
\midrule
  \texttt{opencode} & 66.2\,$\pm$\,3.9 & --- & --- & 66.6\,$\pm$\,2.2 & 14 & 5,051 \\
  \texttt{qwen-coder} & 62.8\,$\pm$\,4.0 & $-3.4$ & 0.46 & 66.9\,$\pm$\,2.2 & 14 & 9,524 \\
  \texttt{mini-swe-agent} & 66.9\,$\pm$\,3.9 & $+0.7$ & 1.00 & 66.8\,$\pm$\,2.2 & 52 & 21,876 \\
\bottomrule
\end{tabular}

\end{table}

\paragraph{Test-time compute buys neither better verdicts nor better-documented ones.}  We compare three explicit levels on one model over the same 148 cases (Table~\ref{tab:ablation-effort}). Compute rises substantially, 38\% from \texttt{low} to \texttt{medium} and 72\% end to end, as median within-case ratios rather than ratios of medians. Quality does not follow. Accuracy spans 0.7 points with every pairwise difference inside 0.1 standard errors and both exact McNemar tests at $p=1.00$. Process adherence does rise monotonically across the three levels, by 4.7 points end to end, but at 1.5\,SE$_d$ that too is inside noise --- the ordering is suggestive and the separation is not established. Additional test-time compute buys neither better verdicts nor a better-documented method, which is what one expects if the binding constraint is retrieval elicitation and label discrimination rather than reasoning depth. One limit is that the trajectories expose no reasoning-token count, so completion tokens are the only available compute proxy and we cannot say \emph{where} the extra compute was spent.

\begin{table}[htbp]
\centering
\caption{\textbf{Test-time compute buys compute, not accuracy.} Opus 4.8 on Claude Code, three explicit reasoning-effort levels over the same 148 development tasks. Completion tokens rise 38\% to \texttt{medium} and 72\% end to end, but no pairwise accuracy difference exceeds 0.1\,SE$_d$ (exact McNemar $p\geq1.00$) and no process difference exceeds 1.5\,SE$_d$. Reasoning depth is not the binding constraint on this benchmark.}
\label{tab:ablation-effort}
\small
\setlength{\tabcolsep}{5pt}
\begin{tabular}{lcccccc}
\toprule
\textbf{Effort} & \textbf{Acc \%} & $\Delta$ & $p$ & \textbf{Proc\%} & \textbf{Comp.\ tok} & $\Delta$\,tok \\
\midrule
  \texttt{low} & 71.6\,$\pm$\,3.7 & --- & --- & 65.0\,$\pm$\,2.2 & 7,332 & --- \\
  \texttt{medium} & 72.3\,$\pm$\,3.7 & $+0.7$ & 1.00 & 68.3\,$\pm$\,2.1 & 10,268 & $+38$\% \\
  \texttt{high} & 72.3\,$\pm$\,3.7 & $+0.7$ & 1.00 & 69.7\,$\pm$\,2.2 & 12,420 & $+72$\% \\
\bottomrule
\end{tabular}
\end{table}

\subsection{Benchmark Validation: Judge Reliability}
\label{sec:judge-validation}
The process score and the reference labels both feed the rankings, so we validate each. The reference labels are validated against independent clinician judgment in Section~\ref{sec:clinical-calibration} (91\% agreement on the calibration sample, $\kappa=0.87$); here we validate the deterministic and LLM-judged components of patient-evidence grounding, process adherence, and policy grounding.

\paragraph{Judge reliability and self-consistency.} The process rubric is LLM-adjudicated, so we measure whether the grade is stable across judges rather than assuming it. A five-judge panel grades every trajectory over the full rubric with three replicates each. We report inter-judge agreement primarily as the mean pairwise disagreement. It flags 16\% of criteria as contested, with the judges self-consistent across replicates and statistically interchangeable on the aggregate score (companion Fleiss' $\kappa \approx 0.65$). The most-contested criteria (for example, a hidden antibiotic de-escalation window and an era-dependent heart-failure therapy) are exactly the cases an independent expert audit flagged as under-specified, so judge disagreement acts as an automatic detector of ambiguous criteria, which are then routed back for expert revision.

\paragraph{Grounding-metric determinism and judge reliability.} The four patient-evidence grounding metrics divide by construction, and each is validated according to what it produces. $P_{\mathrm{record}}$ and $R_{\mathrm{retrieval}}$ are computed by rule from the trace and involve no judge at all: recomputing them over a complete 750-task run returns every per-trial value unchanged. The other two are LLM-judged, so we grade the same reports twice and compare the two passes. $R_{\mathrm{finding}}$ asks the judge one covered$/$not-covered question per rubric category. Because those categories are fixed in advance, the two passes can be compared question by question: over 200 reports, 50 per harness family, they agree on 97.9\% of 1{,}331 category decisions (Cohen's $\kappa=0.95$), 175 of the 200 reports receive an identical score, and no family average moves by more than 0.015. $P_{\mathrm{claim}}$ gives no such fixed question list, because the judge must first break the report into atomic claims and it does not divide it the same way twice; there is nothing to align, so we report how far the score itself moves, which is at most 0.02 and in no consistent direction. We therefore treat the two rule-based metrics as exact, $R_{\mathrm{finding}}$ as reproducible, and differences of 0.02 or less on $P_{\mathrm{claim}}$ as noise.

\paragraph{Policy-judge reliability and self-consistency.} Policy grounding is mostly deterministic: citation resolution and governing-document matching are computed by rule, while only citation support requires LLM judgment. Accordingly, the two rule-based components reproduce exactly, and the LLM-judged support component is stable across judges (companion Fleiss' $\kappa\approx0.86$, within-judge noise 0.023--0.025 on a $0,1$ scale). The production judge is therefore representative rather than unusually strict or lenient. The residual variation is systematic: judges sometimes under-credit near-verbatim citations, making citation-support estimates conservative lower bounds, and we find no evidence of own-family favoritism. Defining governing-document identification through expert-specified document sets rather than LLM judgment removes a major source of ambiguity at the rubric-design stage. This supports treating the policy-grounding metrics as meaningful measurements rather than judge noise.

\paragraph{Judge and label independence.} GPT-5.5 serves three roles: one of three systems in the reference-verdict ensemble, an evaluated system through Codex, and the production judge for the semantically adjudicated metrics. Verdict accuracy involves no judge, being a deterministic four-class match against the reference labels. The labels are bounded by the ensemble and by clinician review: GPT-5.5 casts one of three votes, redundant on the 556 of 750 cases the ensemble decided unanimously, and the reference agrees with each clinician more closely than the two clinicians agree with each other (Section~\ref{sec:clinical-calibration}). Process adherence and policy support are graded by multi-judge panels (Section~\ref{sec:judge-validation}). The remaining exposure is $P_{\mathrm{claim}}$ and $R_{\mathrm{finding}}$, which the production judge grades alone and which regrading validates for stability rather than bias. Codex leads $P_{\mathrm{claim}}$, but GPT-5.5 ranks fourth of its five systems, so the gap does not track the judge's own model. Replication with an independent judge remains the direct test, left for future work.
\section{Conclusion}
\label{sec:conclusion}

\benchmarkname{} reframes clinical-agent evaluation around the capabilities
required for defensible retrospective clinical audit: planning evidence
retrieval across a longitudinal EHR, grounding clinical claims in traceable
patient evidence, applying governing standards, following an observable
and clinically defensible investigation process, and deferring when the
available record cannot support a unique conclusion. Building on the
trace-level clinical auditing introduced by Hager et
al.~\cite{hager2024clinicaldecision}, \benchmarkname{} brings these dimensions
together within a common patient-level adjudication framework.

The benchmark comprises 25 clinician-authored and validated scenarios
instantiated as 750 patient cases over MIMIC-IV. Each run is
evaluated against a case-level reference verdict produced through
independent multi-model adjudication and calibrated against blinded
Clinical Board review on a stratified subset. Beyond final-verdict
correctness, the benchmark evaluates calibrated abstention,
patient-evidence grounding, process adherence, policy grounding,
repeated-run reliability, and resources.

The evaluation exposes two deployment-relevant failure modes that raw
accuracy obscures. First, every evaluated system under-abstains, returning
a definitive verdict on cases for which the reference standard requires
deferral because evidence is missing or the record remains medically
ambiguous. Second, defect-free accuracy falls by 4.8--14.8 percentage
points relative to report-present accuracy and changes the system ranking.
For the most affected system, roughly one in five report-present correct verdicts is reached through a prohibited shortcut. These results show that
a correct final answer does not by itself establish that an agent performed
a defensible clinical investigation.

\paragraph{Scope and limitations.}
\benchmarkname{} trades per-scenario case volume for breadth of workflow
coverage and depth of annotation. It spans 25 scenarios across 14 medical
specialties and ten reasoning capabilities, but contains only 30 cases per
scenario and 750 cases in total, compared with the 2,400 cases concentrated
on four abdominal pathologies in MIMIC-CDM~\cite{hager2024clinicaldecision}.
Its contribution is therefore not greater sample depth within a single
diagnostic family, but the density of its scenario and case
specifications: explicit four-way adjudication criteria, required-evidence
and grounding requirements, weighted \textsc{must-do}/\textsc{must-not}
process rubrics, and clinician-calibrated reference verdicts. Expanding the
number of cases per scenario while preserving this annotation depth is an
important direction for future releases.

The current study also has several important limitations. The benchmark is
derived from a single de-identified academic medical-center dataset, and
its deliberately stratified case distribution should not be interpreted as
clinical prevalence. Reference verdicts are produced by a model-assisted
adjudication procedure and calibrated against clinician review on a subset,
rather than established through exhaustive independent clinician review of
all 750 cases. Clinical Board review in this study calibrates the scenario
specifications and reference verdicts and should not be interpreted as a
direct human-versus-agent performance comparison.

Although developed for clinical audit, the benchmark's central design
principles, including governed and logged tool access, explicit evidentiary
requirements, calibrated abstention, and separate scoring of outcome and
observable process, apply more broadly to other high-stakes domains in
which decisions must be transparent, reproducible, and auditable, such as finance, cybersecurity, and law.

\paragraph{Open-source release and MedHELM integration.} We will release CliniCARE-Bench in two tiers. The non-patient-specific artifacts (scenario
specifications, evaluation framework, harness and metric code, and judge prompts) will
be released openly under a permissive license. The patient-linked artifacts, which are
derived from MIMIC-IV, will be made available on PhysioNet under its credentialed access
policy and data use agreement. We are integrating the benchmark into the MedHELM
evaluation harness \cite{bedi2025medhelm} and, with Pacific AI, upstreaming it alongside
related efforts such as PhysicianBench \cite{liu2026physicianbench} and HealthAdminBench
\cite{bedi2026healthadminbench}, so that its process-aware, grounding- and
abstention-scored cases are usable across a broader open medical-evaluation ecosystem.

\section*{Author Contributions}
Y.~Xue conceived the project, defined the vision and benchmark direction,
led the manuscript effort, and supervised the work.
V.~Chatrath led benchmark execution, including the Clinical Board network and experiments, and served as primary manuscript lead.
B.~Zhu served as healthcare lead and technical lead and was a primary developer.
G.~Pu served as technical lead and was a primary developer.
J.~Fan led evidence-grounding and metrics ideation and implementation.
A.~Shanker led policy-grounding ideation, and V.~Ursekar led process rubric implementation and judge calibration.
A.~Sharma contributed to evidence grounding and the Clinical Board network and created the MIMIC environment.
J.~Qin provided domain expertise and experimental ideation.
K.~Han contributed medical domain expertise, and related work.
S.~D.~Tiwari and S.~Dan contributed to medical case release and environment setup.
Y.~Li and V.~Kalmath contributed to manuscript iteration.
D.~Y.~Zhang oversaw the environment release.
Z.~Doctor contributed medical domain expertise.
Z. Yin served as a research advisor in machine learning and AI for healthcare, guiding clinical agent evaluation design and contributing to the drafting and review of the related-work section. 
C.~Wang served as research advisor for benchmark design, and N.~Shah provided senior clinical and research guidance, including benchmark positioning and connections to the broader medical-AI evaluation community.
All authors contributed to the writing, reviewed and approved the final manuscript.


\clearpage

\bibliographystyle{unsrtnat}
\bibliography{references}

\clearpage
\appendix

\section{Terminology and Units of Evaluation}
\label{app:terminology}

To distinguish the benchmark's clinical specifications, patient-level
instances, and experimental executions, we use the following terminology
throughout the paper.

\begin{table}[h]
\centering
\small
\begin{tabular}{p{2.2cm}p{10.5cm}}
\toprule
\textbf{Term} & \textbf{Definition} \\
\midrule

\textbf{Scenario} &
A reusable, clinician-authored evaluation specification comprising a
clinical-query template, operational adjudication criteria, evidence and
policy-grounding requirements, and a process rubric. The benchmark contains
25 scenarios. \\

\textbf{Case} &
A patient-specific instantiation of a scenario over a particular MIMIC-IV
record and, where applicable, a specific encounter or clinical event.
Each scenario is instantiated on 30 cases, yielding 750 cases in total. \\

\textbf{Run} &
One execution of an evaluated system on one case. A run produces a tool-call
trajectory, retrieved evidence, computations or artifacts, and a final
report containing the system's verdict and supporting citations. \\

\textbf{Attempt} &
The repetition index when the same system is independently run more than
once on the same case. Repeated attempts are used to measure consistency,
including $\mathrm{pass}\mbox{\textasciicircum}k$ and $\mathrm{avg}@k$. \\

\textbf{Cohort} &
A defined collection of cases used for analysis, such as the 30 cases
associated with one scenario, the 148-case development subset, or the full
750-case benchmark. \\

\textbf{System} &
The complete evaluated configuration, including the model, agent harness,
prompting or scaffolding, and available tools. \\

\bottomrule
\end{tabular}
\caption{Terminology and units of evaluation used in \benchmarkname{}.}
\label{tab:terminology}
\end{table}

\newpage
\section{Additional CliniCARE-Bench scenarios}
\label{app:cases}

We detail four representative clinical scenarios below. Each states
its clinical query, the four-way adjudication criteria
(\textit{Yes} / \textit{No} / \textit{Indeterminate: Lack of Data} /
\textit{Indeterminate: Medically Ambiguous}), the evidence a grounded
answer must surface, and the grounding requirement.

\paragraph{Post-CT Acute Kidney Injury.}
This scenario asks whether an ICU patient developed acute kidney injury
within 48 hours after a CT scan, applying the published KDIGO criteria
while explicitly avoiding any inference that the CT or contrast caused
the injury.

\begin{itemize}
    \item \textbf{Clinical Query:} ``For ICU patient \texttt{<subject\_id, hadm\_id>} who underwent a CT this admission, did the patient develop AKI within 48\,h after the CT? Set the baseline creatinine hierarchically---the lowest value in the 3--12\,months before admission; else the admission creatinine (the patient may already present in AKI); else, if CKD is documented, an MDRD estimate assuming eGFR 75\,mL/min/1.73\,m$^2$. Apply the KDIGO serum-creatinine criteria and frame the result as AKI following CT, not as contrast-caused injury; where serum creatinine does not reflect the patient's own kidney function, say the question does not resolve rather than applying the thresholds.''
    \item \textbf{Adjudication Criteria:}
    \begin{itemize}
        \item \textit{Yes:} A baseline is set by the hierarchy (the lowest creatinine in the 3--12\,months pre-admission; else the admission creatinine; else an MDRD estimate where CKD is documented); at least one creatinine value falls $\leq 48$\,h post-CT, with all post-CT values searched for the peak (not only the first); a post-CT value meets a KDIGO threshold (absolute rise $\geq 0.3$\,mg/dL above baseline, or $\geq 1.5\times$ baseline); and the patient is not on chronic dialysis or ESRD.
        \item \textit{No:} A baseline can be established by the hierarchy and at least two creatinine values fall within 48\,h post-CT (adequate sampling); no post-CT value meets either KDIGO threshold (absolute rise $\geq 0.3$\,mg/dL or $\geq 1.5\times$ baseline); and the patient is not on chronic renal replacement therapy and has no ESRD or CKD stage $\geq 4$.
        \item \textit{Indeterminate: Lack of Data:} No creatinine is available to set any tier of the baseline hierarchy (no 3--12\,month prior value, no admission creatinine, and no CKD documentation supporting an MDRD estimate), or none falls within 48\,h post-CT; CT timing cannot be resolved (order time versus actual scan time, or multiple CTs prevent a unique ``first qualifying'' CT); no CT can be identified during the admission, so the premise is unmet and the case is out of scope; or post-CT sampling is insufficient ($<2$ creatinine values) to confirm the peak.
        \item \textit{Indeterminate: Medically Ambiguous:} Chronic dialysis or ESRD (ICD N18.6 / Z99.2 / 585.6 / V45.11, or any hemodialysis order), so the KDIGO acute thresholds do not apply; or a labile baseline---the candidate prior creatinines (3--12\,months before admission) fluctuate materially, so a clean reference cannot be set and a $+0.3$\,mg/dL rise cannot be confidently attributed.
    \end{itemize}
    \item \textbf{Required Evidence:} Creatinine timeline from the baseline source (the lowest value in the 3--12\,months pre-admission; else the admission value; else the CKD/MDRD basis) through 48\,h post-CT, showing every value; ESRD/dialysis status; and the CT \emph{scan} time distinguished from the order time (delays), confirmed via the radiology report where possible.
    \item \textbf{Required Grounding:} The specific MIMIC-IV record or record span from which the decisive finding was derived.
\end{itemize}

\paragraph{Advanced Cross-Sectional Imaging Receipt.}
This scenario asks whether an ICU patient received advanced
cross-sectional imaging---a CT or MRI---at any point during a specific
ICU stay, resolving the question from the order stream confirmed against
the radiology report rather than from an order alone.

\begin{itemize}
    \item \textbf{Clinical Query:} ``For ICU stay \texttt{<subject\_id, hadm\_id, stay\_id>}, did the patient receive advanced cross-sectional imaging (a CT or MRI) at any point during this ICU stay window \texttt{[intime, outtime]}? Decide how to identify imaging events and cite the evidence; an order alone does not settle the question where a confirming radiology report is available, and for a multi-stay admission attribute each order to a single stay before answering.''
    \item \textbf{Adjudication Criteria:}
    \begin{itemize}
        \item \textit{Yes:} At least one POE radiology order (\texttt{order\_type}=``Radiology'') of a CT/MRI-class subtype (``CT Scan'', ``MRI'', or ``Cross-Sectional Interventional Radiology''), with \texttt{ordertime} inside the ICU stay window and a non-discontinued status, together with a radiology report confirming the study was actually performed. An order with no confirming report is not sufficient for \textit{Yes}.
        \item \textit{No:} Within the ICU stay window the patient received only bedside or non-cross-sectional imaging (``General Xray'', ``Ultrasound'', ``Noninvasive Vascular'', or a TEE echo order rather than a radiology CT/MRI), with no CT or MRI; or no radiology order of any kind falls within the window while the order stream for the stay is present and complete, so the absence is a true negative rather than missing data.
        \item \textit{Indeterminate: Lack of Data:} A CT/MRI order was cancelled or discontinued with no confirming report; a multi-ICU-stay admission where the order cannot be attributed to this specific stay; or no ICU stay window is recorded for the patient, so the premise cannot be evaluated.
        \item \textit{Indeterminate: Medically Ambiguous:} The order subtype is genuinely ambiguous as to whether it denotes a cross-sectional study.
    \end{itemize}
    \item \textbf{Required Evidence:} The POE radiology orders (subtype, \texttt{ordertime}, order status); the ICU stay window (\texttt{icustays.intime} / \texttt{outtime} / \texttt{stay\_id}); the confirming radiology report text and \texttt{charttime}; and, for a multi-stay admission, the attribution of each order to a single stay.
    \item \textbf{Required Grounding:} The specific MIMIC-IV record or record span from which the decisive finding was derived.
\end{itemize}

\paragraph{Medication Reconciliation Discrepancy.}
This scenario asks whether an admission contains a clinically significant
medication-reconciliation discrepancy across the three medication
sources---home medications at arrival, the inpatient record, and the
discharge list---distinguishing an unintended discrepancy from a
documented intentional change.

\begin{itemize}
    \item \textbf{Clinical Query:} ``For admission \texttt{<subject\_id, hadm\_id>} (with ED stay \texttt{<stay\_id>}), reconcile the discharge medication list against the home medications recorded at arrival and the medications administered in the last 24\,h. Are there clinically significant reconciliation discrepancies? List each discrepancy with its type and significance; a difference whose intent---a deliberate change versus an unintended omission---cannot be established from the record should be escalated rather than called a discrepancy.''
    \item \textbf{Adjudication Criteria:}
    \begin{itemize}
        \item \textit{Yes:} At least one high-significance unintended discrepancy---an omission of a chronic medication, a home medication held and never restarted, a time-limited course with a missing duration, or an undocumented dose, frequency, or route change---for which no deliberate clinical rationale is documented.
        \item \textit{No:} All three sources reconcile, and every difference is either trivial or a documented intentional change.
        \item \textit{Indeterminate: Lack of Data:} There is no ED encounter (so no home-medication reconciliation list) or no structured discharge list, so the three-way reconciliation cannot be performed.
        \item \textit{Indeterminate: Medically Ambiguous:} A difference exists but its intent---a deliberate change versus an unintended omission---cannot be determined from the record, so that item is escalated rather than adjudicated as a discrepancy.
    \end{itemize}
    \item \textbf{Required Evidence:} Home medications at ED arrival (MIMIC-IV-ED \texttt{medrecon}); active inpatient orders and the last-24\,h administrations (\texttt{prescriptions}, \texttt{emar}, \texttt{pharmacy}); and the discharge medication list from both the discharge note and \texttt{prescriptions}---compared home~$\rightarrow$~inpatient~$\rightarrow$~discharge, with the last-24\,h window measured to \texttt{dischtime}.
    \item \textbf{Required Grounding:} The specific MIMIC-IV record or record span from which the decisive finding was derived.
\end{itemize}

\paragraph{Longitudinal Renal-Function Trajectory.}
This scenario asks whether a patient's renal function is on a worsening
trajectory across all recorded admissions, assessed by estimated GFR
rather than serum creatinine alone, since creatinine understates GFR
decline as muscle mass falls.

\begin{itemize}
    \item \textbf{Clinical Query:} ``For patient \texttt{<subject\_id>} with multiple admissions, is renal function on a worsening trajectory (versus stable or improving) across their course? Assess by estimated GFR (via a validated equation such as the 2021 CKD-EPI, or MDRD / Cockcroft--Gault) rather than serum creatinine alone, since creatinine understates GFR decline as muscle mass falls; cite the specific admissions, the values relied on, and the inflection points.''
    \item \textbf{Adjudication Criteria:}
    \begin{itemize}
        \item \textit{Yes:} A sustained decline in per-admission baseline eGFR across $\geq 2$ admissions beyond a defined delta, or the onset of kidney replacement therapy (hemodialysis, peritoneal dialysis, or transplantation), or documented CKD-stage progression with concordant labs (or, where available across $\geq 2$ admissions, a sustained shift in cystatin-C beyond the delta). A declining-eGFR trajectory is robust to the muscle-mass confound.
        \item \textit{No:} Per-admission baseline eGFR is flat or improving across admissions; a discrete, reversible AKI that recovers to baseline does not count as worsening. Where both this criterion and the medical-ambiguity clause below could fit---flat creatinine in an older, frail patient---this \textit{No} criterion governs.
        \item \textit{Indeterminate: Lack of Data:} Fewer than two admissions, or creatinine sampling too sparse to construct per-admission baselines; or the trajectory is dominated by a single acute, reversible event with no clear baseline.
        \item \textit{Indeterminate: Medically Ambiguous:} Flat or mildly rising creatinine in an older, frail, or long-course patient, where muscle loss means stable creatinine cannot exclude real GFR decline---especially with coded CKD and no cystatin-C or muscle-corrected eGFR available (the usual case in MIMIC-IV, 2008--2019)---provided the per-admission eGFR trajectory is not itself flat or improving; or lab-versus-ICD discordance, where codes assert CKD or progression but creatinine is mild or flat and a human read is needed.
    \end{itemize}
    \item \textbf{Required Evidence:} Per-admission baseline, peak, and discharge creatinine with dates; reno-active medication changes; discharge-summary spans for the inflection points; and any weight/BMI trend and any cystatin-C or eGFR values---with an explicit note when these are absent.
    \item \textbf{Required Grounding:} The specific MIMIC-IV record or record span from which the decisive finding was derived.
\end{itemize}

\newpage

\subsection{Representative Scenario Selection}
\label{app:task_coverage}

Table~\ref{tab:secondary_case_selection} highlights a representative subset
to convey the breadth of the suite. Scenario numbers are consistent with the index
in Figure~\ref{fig:coverage}.

\begin{table}[htbp]
\centering
\caption{\textbf{Representative Scenario Selection (Suite Overview)}}
\label{tab:secondary_case_selection}
\small
\renewcommand{\arraystretch}{1.25}
\setlength{\tabcolsep}{6pt}
\begin{tabular}{@{}c l >{\raggedright\arraybackslash}p{9.4cm}@{}}
\toprule
\textbf{Scenario} & \textbf{Short Title} & \textbf{Brief Core Objective} \\
\midrule
\textbf{8}  & Antibiotic Stewardship      & Evaluate whether empiric therapy covered the cultured organism's susceptibilities and was narrowed within 48\,h once a susceptible narrower-spectrum option was available. \\
\addlinespace
\textbf{10} & DKA Resolution Timeline      & Reconstruct the diabetic-ketoacidosis resolution timeline and flag protocol deviations (e.g., dextrose not started when glucose dropped below 200\,mg/dL, or no SC-insulin overlap before drip discontinuation). \\
\addlinespace
\textbf{12} & Hyperacute Stroke Audit      & Compute door-to-needle and door-to-puncture intervals against AHA targets and extract documented justifications for withholding thrombolysis. \\
\addlinespace
\textbf{13} & Hepatic Encephalopathy       & Audit narrative and structured data to confirm or rule out the standard HE precipitants (e.g., active infection, GI hemorrhage, electrolyte derangement, lactulose non-adherence). \\
\addlinespace
\textbf{15} & ICU Delirium Screening       & Confirm CAM-ICU (or equivalent) screening was documented per shift and, where delirium was present, that an ABCDEF bundle response was applied and deliriogenic agents (benzodiazepines, anticholinergics, meperidine) were avoided or clinically justified. \\
\addlinespace
\textbf{17} & Anticoagulation Reversal     & Audit whether reversal matched current guidelines for the specific anticoagulant and bleed (e.g., 4-factor PCC over FFP for warfarin major bleed; agent-specific reversal for DOACs). \\
\addlinespace
\textbf{19} & Hyperkalemia Sequence        & Verify adherence to the safety sequence (calcium stabilization before intracellular shifting and elimination) and measure time-to-calcium for critical potassium values ($\ge 6.5$\,mEq/L). \\
\addlinespace
\textbf{22} & End-of-Life Concordance      & Audit the final 72 hours of life to determine whether intensity of intervention (ventilation, pressors, dialysis, procedures) was concordant with documented goals of care. \\
\addlinespace
\textbf{24} & Frequent-Admitter Synthesis  & Synthesize a single patient's recurrent admissions to extract recurring diagnoses and $\ge 1$ documented intervenable driver (e.g., medication non-adherence, substance use, housing instability). \\
\bottomrule
\end{tabular}
\end{table}

\newpage
\section{Per-class precision and recall}
\label{app:per-class}

\begin{table}[htbp]
\centering
\caption{\textbf{Per-class F1, with precision and recall for the two \textit{Indeterminate}
classes.} The two rightmost cells read P\,$/$\,R. Denominators follow the cohort's natural label
distribution, not the development subset's balanced one, so cells are not comparable with
Table~\ref{tab:dev-perclass}.}
\label{tab:main750-perclass}
\small
\setlength{\tabcolsep}{4pt}
\begin{tabular}{lccccccc}
\toprule
 & & \multicolumn{4}{c}{\textbf{F1}} & \multicolumn{2}{c}{\textbf{Precision\,$/$\,Recall \%}} \\
\cmidrule(lr){3-6}\cmidrule(lr){7-8}
\textbf{Model (harness)} & \textbf{n} & Yes & No & lack & amb & lack & amb \\
\midrule
Opus 5 (Claude Code)           & 750 & 0.809 & 0.771 & 0.628 & 0.596 & 60.0\,/\,65.9 & 56.7\,/\,63.0 \\
Sonnet 5 (Claude Code)         & 750 & 0.796 & 0.749 & 0.643 & 0.426 & 61.5\,/\,67.5 & 38.2\,/\,48.1 \\
\addlinespace
GPT-5.6-Sol (Codex)            & 750 & 0.795 & 0.740 & 0.619 & 0.500 & 57.7\,/\,66.7 & 45.5\,/\,55.6 \\
GPT-5.6-Luna (Codex)           & 750 & 0.739 & 0.708 & 0.494 & 0.411 & 45.8\,/\,53.7 & 32.6\,/\,55.6 \\
GPT-5.5 (Codex)                & 750 & 0.830 & 0.762 & 0.605 & 0.600 & 62.6\,/\,58.5 & 65.2\,/\,55.6 \\
GPT-5.4 (Codex)                & 750 & 0.775 & 0.736 & 0.555 & 0.464 & 53.4\,/\,57.7 & 44.8\,/\,48.1 \\
GPT-5.4-mini (Codex)           & 750 & 0.764 & 0.678 & 0.442 & 0.353 & 48.5\,/\,40.7 & 29.3\,/\,44.4 \\
\addlinespace
Gemini-3.6-Flash (Gemini CLI)  & 750 & 0.801 & 0.714 & 0.621 & 0.473 & 66.1\,/\,58.5 & 46.4\,/\,48.1 \\
Gemini-3.5-Flash (Gemini CLI)  & 750 & 0.786 & 0.714 & 0.611 & 0.377 & 66.0\,/\,56.9 & 38.5\,/\,37.0 \\
Gemini-3.1-Pro (Gemini CLI)    & 750 & 0.771 & 0.733 & 0.653 & 0.500 & 56.5\,/\,77.2 & 64.7\,/\,40.7 \\
\addlinespace
DeepSeek-V4-Pro (opencode)     & 750 & 0.752 & 0.695 & 0.556 & 0.444 & 58.6\,/\,52.8 & 38.9\,/\,51.9 \\
DeepSeek-V4-Flash (opencode)   & 750 & 0.800 & 0.705 & 0.500 & 0.340 & 58.1\,/\,43.9 & 34.6\,/\,33.3 \\
GLM-5.2 (opencode)             & 750 & 0.804 & 0.743 & 0.597 & 0.426 & 67.3\,/\,53.7 & 38.2\,/\,48.1 \\
Qwen-3.7-Plus (opencode)       & 750 & 0.737 & 0.678 & 0.438 & 0.463 & 46.4\,/\,41.5 & 34.5\,/\,70.4 \\
MiniMax-M3 (opencode)          & 750 & 0.742 & 0.690 & 0.408 & 0.380 & 52.6\,/\,33.3 & 28.8\,/\,55.6 \\
Kimi-K2.7-Code (opencode)      & 750 & 0.751 & 0.699 & 0.431 & 0.350 & 54.3\,/\,35.8 & 26.4\,/\,51.9 \\
\bottomrule
\end{tabular}
\end{table}

\newpage
\section{Verdict accuracy per scenario}
\label{app:per-scenario}

\begin{table}[htbp]
\centering
\caption{\textbf{Verdict accuracy (\%) per scenario, all 16 systems.} Rows are the 25 clinical
scenarios, sorted by the unweighted mean over systems given in the last column; the best cell in each row
is bold. A cell's denominator is that system's graded runs for that scenario, exactly 30 for every
cell in the grid, so a single cell is a weaker estimate than a row or a column.
Column key: O5\,=\,Opus 5, S5\,=\,Sonnet 5, Sol\,=\,GPT-5.6-Sol, Luna\,=\,GPT-5.6-Luna, 5.5\,=\,GPT-5.5, 5.4\,=\,GPT-5.4, 5.4m\,=\,GPT-5.4-mini, G3.6\,=\,Gemini-3.6-Flash, G3.5\,=\,Gemini-3.5-Flash, G3.1\,=\,Gemini-3.1-Pro, D4P\,=\,DeepSeek-V4-Pro, D4F\,=\,DeepSeek-V4-Flash, GLM\,=\,GLM-5.2, Qwn\,=\,Qwen-3.7-Plus, MM3\,=\,MiniMax-M3, K2.7\,=\,Kimi-K2.7-Code.}
\label{tab:scenario-table}
\small
\setlength{\tabcolsep}{3pt}
\resizebox{\textwidth}{!}{%
\begin{tabular}{lccccccccccccccccc}
\toprule
\textbf{Scenario} & \textbf{O5} & \textbf{S5} & \textbf{Sol} & \textbf{Luna} & \textbf{5.5} & \textbf{5.4} & \textbf{5.4m} & \textbf{G3.6} & \textbf{G3.5} & \textbf{G3.1} & \textbf{D4P} & \textbf{D4F} & \textbf{GLM} & \textbf{Qwn} & \textbf{MM3} & \textbf{K2.7} & \textbf{Mean} \\
\midrule
eol-goal-concordance     & 93 & 93 & \textbf{100} & 97 & 93 & 97 & 93 & 90 & 90 & \textbf{100} & 77 & 93 & 90 & 83 & 87 & 90 & 91.7 \\
overdose-disposition     & 87 & \textbf{97} & 87 & 83 & 90 & 90 & \textbf{97} & \textbf{97} & \textbf{97} & 90 & 93 & \textbf{97} & 90 & 80 & 80 & \textbf{97} & 90.6 \\
renal-trajectory         & 87 & 93 & 90 & 90 & 93 & 90 & \textbf{97} & 83 & 93 & 90 & 90 & 87 & 83 & 87 & 90 & 83 & 89.2 \\
transfusion-audit        & 87 & \textbf{90} & 87 & 83 & \textbf{90} & \textbf{90} & 87 & \textbf{90} & \textbf{90} & \textbf{90} & 87 & \textbf{90} & \textbf{90} & 83 & \textbf{90} & 83 & 87.9 \\
discharge-audit          & \textbf{93} & 77 & 80 & 83 & 90 & 90 & 63 & 80 & 83 & 90 & 77 & 73 & 80 & 80 & 80 & 83 & 81.5 \\
advanced-imaging         & 83 & 87 & 87 & \textbf{90} & 83 & 77 & 67 & 87 & 87 & 77 & 80 & 73 & 87 & 77 & 73 & 77 & 80.6 \\
hfref-gdmt               & 80 & \textbf{87} & 83 & 63 & 83 & 73 & 70 & 83 & 83 & 77 & 80 & 73 & 80 & 73 & 70 & 70 & 76.9 \\
post-ct-aki              & 83 & 83 & 90 & \textbf{93} & 90 & 73 & 80 & 73 & 77 & 73 & 60 & 77 & 67 & 60 & 80 & 67 & 76.7 \\
sepsis-bundle            & 83 & 80 & 83 & 80 & 83 & 77 & 77 & 83 & 80 & \textbf{87} & 67 & 73 & 83 & 67 & 53 & 60 & 76.0 \\
goals-of-care            & 93 & 90 & 90 & 60 & 87 & 83 & 57 & 67 & 70 & \textbf{97} & 87 & 77 & 83 & 63 & 57 & 53 & 75.8 \\
frequent-admitter        & 77 & 70 & 83 & 77 & 77 & 63 & 63 & \textbf{87} & 77 & 83 & 67 & 73 & 83 & 77 & 63 & 67 & 74.2 \\
readmission-30d          & 70 & 70 & 73 & 80 & 80 & 77 & 67 & 77 & 70 & \textbf{83} & 73 & 63 & 73 & 70 & 70 & 60 & 72.3 \\
icu-delirium             & 67 & \textbf{83} & 60 & 63 & 63 & 67 & 67 & 77 & 73 & 63 & 60 & 60 & 77 & 67 & 63 & 80 & 68.1 \\
sepsis-source-abx        & \textbf{80} & 60 & 63 & 43 & 70 & 67 & 73 & \textbf{80} & 67 & 73 & 57 & 77 & 77 & 57 & 67 & 47 & 66.0 \\
med-reconciliation       & \textbf{87} & 70 & 73 & 57 & 60 & 60 & 53 & 60 & 70 & 83 & 73 & 53 & 83 & 57 & 57 & 60 & 66.0 \\
postop-complications     & 53 & 60 & 57 & 63 & 70 & 67 & 67 & 70 & 67 & 67 & 70 & \textbf{77} & 63 & 63 & 67 & 67 & 65.4 \\
pe-workup                & \textbf{77} & 63 & 70 & 63 & 67 & 63 & 50 & 63 & 63 & 67 & 63 & 67 & 57 & 67 & 57 & 57 & 63.3 \\
dka-resolution           & 83 & 57 & 87 & 60 & \textbf{90} & 77 & 23 & 73 & 67 & 60 & 53 & 47 & 70 & 43 & 53 & 63 & 62.9 \\
hyperkalemia-management  & 60 & 57 & 63 & 60 & 70 & 70 & 67 & 47 & 53 & 53 & 67 & 60 & \textbf{80} & 63 & 63 & 67 & 62.5 \\
ams-workup               & 60 & \textbf{80} & 63 & 50 & 67 & 67 & 50 & 70 & 60 & 57 & 60 & 50 & 67 & 63 & 53 & 73 & 61.9 \\
empiric-abx              & 70 & \textbf{73} & 63 & 57 & 70 & 63 & 63 & 63 & 63 & 67 & 63 & 60 & 67 & 50 & 40 & 43 & 61.0 \\
he-precipitant           & 57 & 53 & 60 & 57 & 60 & 57 & 67 & 67 & 63 & 63 & 60 & 67 & 63 & 57 & 53 & \textbf{73} & 61.0 \\
stroke-time-targets      & 67 & 57 & 53 & 50 & 63 & 57 & 53 & 70 & \textbf{73} & 53 & 63 & 67 & 57 & 53 & 67 & 60 & 60.2 \\
anticoag-reversal        & 67 & 70 & 60 & 40 & 60 & 57 & 47 & 67 & 70 & 57 & 53 & \textbf{80} & 50 & 57 & 63 & 40 & 58.5 \\
pneumonia-curb65         & 47 & 47 & 23 & 27 & 53 & 27 & \textbf{63} & 20 & 10 & 17 & 37 & 53 & 40 & 37 & 50 & 43 & 37.1 \\
\bottomrule
\end{tabular}
}
\end{table}

\newpage
\section{Development-subset experiments}
\label{app:dev-subset}

The main text reports the 750-case cohort. This appendix carries the experiments measured on the
148-case \emph{development subset}. The \textit{dev subset} is a frozen subset of 148 of the 750 cases, built by taking up to
two patients per verdict class from each scenario. It over-samples the two rare \textit{Indeterminate}
classes (gold supports \textit{Yes}~50, \textit{No}~45, \textit{lack}~39, \textit{amb}~14), so an
intervention aimed at deferral can move a measurable number of cases. Its case ids are a verified subset of the 750.

\subsection{Sixteen-system comparison and per-class detail}
\label{app:dev-leaderboard}

\paragraph{Sixteen systems, and the dissociations rather than the ranking.} Accuracy on the subset
spans 59.5--71.6 and does not resolve at the top (Table~\ref{tab:dev-leaderboard}): Opus 5 leads at
71.6 with GPT-5.5 0.7 behind and Gemini-3.1-Pro 0.7 behind that, three systems from three harness
families inside 1.4 points --- well under the $\pm$3.7-point standard error at this $n$. The subset is
too small to separate them: of the $\binom{16}{2}=120$ pairs, only 11 reach
$p<0.05$ under two-sided exact McNemar on paired per-case outcomes and \emph{none} survives Holm
correction.

\begin{table}[htbp]
\centering
\caption{\textbf{Sixteen-system comparison, development subset.} \textit{148 cases, single attempt,
clinician-calibrated reference verdicts; all arms complete.} Columns as in Table~\ref{tab:main750}. ECE is expected calibration
error (Eq.~\ref{eq:ece}); it is not comparable across all arms, because stated confidence is
near-degenerate for the Gemini CLI arms --- one emits three distinct values across 148 reports --- so
those figures indicate an absence of graded uncertainty rather than a finer ranking.
Rows are obtained by \emph{slicing} each arm's completed 750-case run to the 148 subset ids, so the
Claude arm here is Opus 5 --- unlike the paired ablations of Section~\ref{sec:ablations}, which reuse
the earlier Opus 4.8 runs.}
\label{tab:dev-leaderboard}
\small
\setlength{\tabcolsep}{4pt}
\resizebox{\textwidth}{!}{%
\begin{tabular}{lccccccccc}
\toprule
\textbf{Model (harness)} & \textbf{n} & \textbf{Acc} & \textbf{Macro-F1} & \textbf{Def-free} & \textbf{Gap} & \textbf{Proc\%} & \textbf{ECE} & \textbf{Over-com} & \textbf{Over-abs} \\
\midrule
Opus 5 (Claude Code)          & 148 & \textbf{71.6} & \textbf{0.703} & 66.9 & \phantom{0}4.7 & \textbf{80.7} & \textbf{0.096} & 35.8 & \phantom{0}9.5 \\
Sonnet 5 (Claude Code)        & 148 & 66.2 & 0.626 & 56.8 & \phantom{0}9.5 & 68.4 & 0.119 & 35.8 & 12.6 \\
\addlinespace
GPT-5.6-Sol (Codex)           & 148 & 68.2 & 0.666 & 62.8 & \phantom{0}5.4 & 77.4 & 0.239 & 37.7 & 11.6 \\
GPT-5.6-Luna (Codex)          & 148 & 61.5 & 0.594 & 55.4 & \phantom{0}6.1 & 73.1 & 0.278 & 39.6 & 16.8 \\
GPT-5.5 (Codex)               & 148 & 70.9 & 0.687 & \textbf{68.2} & \textbf{\phantom{0}2.7} & 74.5 & 0.135 & 47.2 & \phantom{0}7.4 \\
GPT-5.4 (Codex)               & 148 & 68.2 & 0.643 & 62.8 & \phantom{0}5.4 & 72.7 & 0.174 & 35.8 & \phantom{0}8.4 \\
GPT-5.4-mini (Codex)          & 148 & 61.5 & 0.549 & 54.1 & \phantom{0}7.4 & 63.3 & 0.253 & 56.6 & 11.6 \\
\addlinespace
Gemini-3.6-Flash (Gemini CLI) & 148 & 67.6 & 0.649 & 54.7 & 12.8 & 70.2 & 0.302 & 43.4 & \textbf{\phantom{0}5.3} \\
Gemini-3.5-Flash (Gemini CLI) & 148 & 67.6 & 0.626 & 56.8 & 10.8 & 71.4 & 0.301 & 47.2 & \phantom{0}6.3 \\
Gemini-3.1-Pro (Gemini CLI)   & 148 & 70.3 & 0.668 & 60.1 & 10.1 & 56.2 & 0.273 & \textbf{26.4} & \phantom{0}7.4 \\
\addlinespace
DeepSeek-V4-Pro (opencode)    & 148 & 64.9 & 0.613 & 59.5 & \phantom{0}5.4 & 67.4 & 0.228 & 41.5 & \phantom{0}8.4 \\
DeepSeek-V4-Flash (opencode)  & 148 & 63.5 & 0.557 & 55.4 & \phantom{0}8.1 & 66.3 & 0.298 & 56.6 & \phantom{0}6.3 \\
GLM-5.2 (opencode)            & 148 & 66.9 & 0.631 & 58.8 & \phantom{0}8.1 & 74.6 & 0.188 & 45.3 & \phantom{0}9.5 \\
Qwen-3.7-Plus (opencode)      & 148 & 59.5 & 0.565 & 54.1 & \phantom{0}5.4 & 66.1 & 0.240 & 45.3 & 15.8 \\
MiniMax-M3 (opencode)         & 148 & 60.1 & 0.552 & 52.0 & \phantom{0}8.1 & 72.1 & 0.244 & 56.6 & 11.6 \\
Kimi-K2.7-Code (opencode)     & 148 & 60.8 & 0.557 & 48.0 & 12.8 & 66.3 & 0.212 & 60.4 & \phantom{0}9.5 \\
\bottomrule
\end{tabular}%
}
\end{table}

\paragraph{Over-commitment exceeds over-abstention in every system without exception.} Measured
directionally (last two columns of Table~\ref{tab:dev-leaderboard}), over-commitment runs
26.4--60.4\% against 53 deferral cases while over-abstention runs 5.3--16.8\% against 95 definitive
ones, and \emph{all sixteen} systems over-commit at the higher rate --- reproducing
Section~\ref{sec:main-results} on a deliberately different label distribution. The two directions are
not inverses: Gemini-3.6-Flash posts the lowest over-abstention (5.3\%) alongside a high
over-commitment (43.4\%), whereas Gemini-3.1-Pro achieves the lowest over-commitment (26.4\%) without
a corresponding penalty (7.4\%), so a system can improve on one axis without paying on the other.

\paragraph{Ambiguity elicitation responds to intervention; benchmark difficulty does not.} Revising
the ambiguity guidance in the case prompts raised pooled recall on the \textit{amb} class roughly
threefold, improving in every arm measured with none regressing, while accuracy over the same change
stayed flat and no per-arm shift survived correction. That measurement was taken on the pre-revision
prompt build, before the label revision described in Section~\ref{sec:ablations}, so we report its
direction and withhold the figures rather than invite comparison with the tables here. Elicitation
improved; the benchmark's difficulty did not.

\paragraph{Per-class detail.} Table~\ref{tab:dev-perclass} gives per-class F1 across the sixteen
arms, alongside the macro average that the leaderboard reports. \textit{lack} F1 spreads the field
more widely than accuracy does --- 0.62--0.73 for the strongest arms against 0.35--0.47 for the
weakest --- on a safety-critical class. It does \emph{not}, however, separate the tiers: GPT-5.4-mini
reaches only 0.441, below both GLM-5.2 (0.603) and DeepSeek-V4-Pro (0.597), so the vendor and
open-weight ranges overlap here as they do on accuracy. At an \textit{amb} support of 14 one case moves
a cell by 7.1 points, so that column supports no ranking.

\begin{table}[htbp]
\centering
\caption{\textbf{Per-class F1, development subset.} All arms $n=148$; gold supports \textit{Yes} 50,
\textit{No} 45, \textit{lack} 39, \textit{amb} 14. Macro is the unweighted mean of the four, so the two
rare \textit{Indeterminate} classes carry equal weight with the common ones. Ordered as in
Table~\ref{tab:dev-leaderboard}.}
\label{tab:dev-perclass}
\small
\setlength{\tabcolsep}{4pt}
\begin{tabular}{lccccc}
\toprule
\textbf{Model (harness)} & \textbf{Yes} & \textbf{No} & \textbf{lack} & \textbf{amb} & \textbf{Macro} \\
\midrule
Opus 5 (Claude Code)           & 0.762 & \textbf{0.716} & 0.667 & \textbf{0.667} & \textbf{0.703} \\
Sonnet 5 (Claude Code)         & 0.729 & 0.644 & 0.649 & 0.480 & 0.626 \\
\addlinespace
GPT-5.6-Sol (Codex)            & 0.772 & 0.633 & 0.649 & 0.609 & 0.666 \\
GPT-5.6-Luna (Codex)           & 0.688 & 0.627 & 0.541 & 0.519 & 0.594 \\
GPT-5.5 (Codex)                & \textbf{0.808} & 0.712 & 0.562 & \textbf{0.667} & 0.687 \\
GPT-5.4 (Codex)                & 0.752 & 0.700 & 0.620 & 0.500 & 0.643 \\
GPT-5.4-mini (Codex)           & 0.707 & 0.691 & 0.441 & 0.357 & 0.549 \\
\addlinespace
Gemini-3.6-Flash (Gemini CLI)  & 0.766 & 0.653 & 0.594 & 0.583 & 0.649 \\
Gemini-3.5-Flash (Gemini CLI)  & 0.729 & 0.706 & 0.615 & 0.455 & 0.626 \\
Gemini-3.1-Pro (Gemini CLI)    & 0.738 & 0.681 & \textbf{0.725} & 0.526 & 0.668 \\
\addlinespace
DeepSeek-V4-Pro (opencode)     & 0.722 & 0.653 & 0.597 & 0.480 & 0.613 \\
DeepSeek-V4-Flash (opencode)   & 0.765 & 0.646 & 0.483 & 0.333 & 0.557 \\
GLM-5.2 (opencode)             & 0.734 & 0.688 & 0.603 & 0.500 & 0.631 \\
Qwen-3.7-Plus (opencode)       & 0.713 & 0.592 & 0.469 & 0.485 & 0.565 \\
MiniMax-M3 (opencode)          & 0.692 & 0.667 & 0.377 & 0.471 & 0.552 \\
Kimi-K2.7-Code (opencode)      & 0.738 & 0.636 & 0.353 & 0.500 & 0.557 \\
\bottomrule
\end{tabular}
\end{table}

\subsection{Policy grounding on the development subset}
\label{app:policy-dev}

Table~\ref{tab:policy} reports the policy axes on the development subset, where every arm is complete at
$n=148$. It is separated from the cohort result for the reasons above; Table~\ref{tab:main750-policy}
carries the figures the main text uses.

\begin{table}[htbp]
\centering
\caption{\textbf{Policy grounding, development subset.} Document-set precision and recall are
deterministic set membership against the expert-specified governing set; support is the LLM-judged
check that a cited passage entails its claim. \emph{Fab.} counts citations resolving
to no corpus document. Citation resolution (0.95--1.00 for every arm) and F1 (determined by the two
columns beside it) are omitted as non-discriminating.}
\label{tab:policy}
\small
\begin{tabular}{lcccr}
\toprule
\textbf{Model (harness)} & \textbf{Support} & \textbf{Doc-P} & \textbf{Doc-R} & \textbf{Fab.} \\
\midrule
Opus 5 (Claude Code)           & 0.551 & 0.753 & \textbf{0.686} & \phantom{0}0 \\
Sonnet 5 (Claude Code)         & 0.566 & 0.802 & 0.421 & \phantom{0}6 \\
\addlinespace
GPT-5.6-Sol (Codex)            & 0.732 & 0.820 & 0.575 & \phantom{0}0 \\
GPT-5.6-Luna (Codex)           & \textbf{0.834} & 0.804 & 0.498 & \phantom{0}3 \\
GPT-5.5 (Codex)                & 0.744 & 0.764 & 0.562 & \phantom{0}0 \\
GPT-5.4 (Codex)                & 0.632 & 0.779 & 0.531 & \phantom{0}0 \\
GPT-5.4-mini (Codex)           & 0.634 & 0.773 & 0.461 & \phantom{0}0 \\
\addlinespace
Gemini-3.6-Flash (Gemini CLI)  & 0.454 & 0.771 & 0.635 & 16 \\
Gemini-3.5-Flash (Gemini CLI)  & 0.474 & 0.737 & 0.578 & 71 \\
Gemini-3.1-Pro (Gemini CLI)    & 0.652 & 0.792 & 0.396 & \phantom{0}0 \\
\addlinespace
DeepSeek-V4-Pro (opencode)     & 0.605 & 0.861 & 0.517 & \phantom{0}3 \\
DeepSeek-V4-Flash (opencode)   & 0.585 & 0.842 & 0.418 & 25 \\
GLM-5.2 (opencode)             & 0.569 & 0.812 & 0.562 & 16 \\
Qwen-3.7-Plus (opencode)       & 0.582 & \textbf{0.870} & 0.401 & \phantom{0}1 \\
MiniMax-M3 (opencode)          & 0.453 & 0.799 & 0.543 & 29 \\
Kimi-K2.7-Code (opencode)      & 0.651 & 0.806 & 0.532 & \phantom{0}0 \\
\bottomrule
\end{tabular}
\end{table}

\end{document}